\documentclass[10pt,journal,compsoc]{IEEEtran}
\usepackage{amsmath}
\usepackage{bbding}
\usepackage{amssymb}

\usepackage{epsfig}
\usepackage{graphicx}
\usepackage[utf8]{inputenc}
\usepackage[T1]{fontenc}
\usepackage{url}
\usepackage{nicefrac}
\usepackage{microtype}
\usepackage{lipsum}
\usepackage{graphicx}
\usepackage{ragged2e}
\usepackage{booktabs}
\usepackage{amsfonts}
\usepackage{verbatim}
\usepackage{makecell}
\usepackage{multirow}

\usepackage{listings}
\usepackage{color}
\usepackage[dvipsnames]{xcolor}
\definecolor{dkgreen}{rgb}{0,0.6,0}
\definecolor{gray}{rgb}{0.5,0.5,0.5}
\definecolor{mauve}{rgb}{0.58,0,0.82}
\usepackage{pifont}
\usepackage{bm}
\usepackage{caption}
\usepackage{booktabs,subcaption,amsfonts,dcolumn}
\usepackage[ruled,vlined]{algorithm2e}
\usepackage[switch]{lineno}
\usepackage[pagebackref=false,breaklinks=false,colorlinks,citecolor=blue,bookmarks=false]{hyperref}

\usepackage{fancyhdr}
\usepackage{colortbl}
\definecolor{mygray}{gray}{.93}
\usepackage{arydshln}

\begin{document}

\title{Towards Non-Exemplar Semi-Supervised Class-Incremental Learning}

\author{Wenzhuo Liu,
		Fei Zhu,
		Cheng-Lin Liu~\IEEEmembership{Fellow,~IEEE}
	\IEEEcompsocitemizethanks{\IEEEcompsocthanksitem Wenzhuo Liu and Cheng-Lin Liu are with the University of Chinese Academy of Sciences, Beijing, P.R. China, and the State Key Laboratory of Multimodal Artificial Intelligence Systems, Institute of Automation, Chinese Academy of Sciences, 95 Zhongguancun East Road, Beijing 100190, P.R. China. \protect
    \IEEEcompsocthanksitem Fei Zhu is with the Centre for Artificial Intelligence and Robotics, Hong Kong Institute of Science and Innovation, Chinese Academy of Sciences, Hong Kong 999077, China. \protect
		\IEEEcompsocthanksitem  Email: liuwenzhuo20@mails.ucas.ac.cn, fei.zhu@cair-cas.org.hk, liucl@nlpr.ia.ac.cn}
}

\markboth{Journal of \LaTeX\ Class Files,~Vol.~14, No.~8, August~2015}%
{Shell \MakeLowercase{\textit{et al.}}: Bare Advanced Demo of IEEEtran.cls for IEEE Computer Society Journals}

\IEEEtitleabstractindextext{%
\begin{abstract}
Deep neural networks perform remarkably well in close-world scenarios. However, novel classes emerged continually in real applications, making it necessary to learn incrementally. Class-incremental learning (CIL) aims to gradually recognize new classes while maintaining the discriminability of old ones. Existing CIL methods have two limitations: a heavy reliance on preserving old data for forgetting mitigation and the need for vast labeled data for knowledge adaptation. To overcome these issues, we propose a non-exemplar semi-supervised CIL framework with contrastive learning and semi-supervised incremental prototype classifier (Semi-IPC). On the one hand, contrastive learning helps the model learn rich representations, easing the trade-off between learning representations of new classes and forgetting that of old classes.
On the other hand, Semi-IPC learns a prototype for each class with unsupervised regularization, enabling the model to incrementally learn from partially labeled new data while maintaining the knowledge of old classes. Experiments on benchmark datasets demonstrate the strong performance of our method: without storing any old samples and only using less than 1\% of labels, Semi-IPC outperforms advanced exemplar-based methods. We hope our work offers new insights for future CIL research. The code will be made publicly available.

\end{abstract}

\begin{IEEEkeywords}
Incremental Learning, Continual Learning, Catastrophic Forgetting, Semi-supervised Learning, Prototype Learning
\end{IEEEkeywords}}

\maketitle

\IEEEdisplaynontitleabstractindextext
\IEEEpeerreviewmaketitle

\ifCLASSOPTIONcompsoc
\IEEEraisesectionheading{\section{Introduction}\label{sec:introduction}}
\else
\section{Introduction}
\label{sec:introduction}

\fi
 
\IEEEPARstart{A}{daptability} to changing environments is a characteristic shared by humans and large primates. This stems from an innate skill that allows them to consistently acquire and enrich the knowledge throughout their lifetime, while preserving previous knowledge. This ability, called incremental learning or continual learning \cite{wixted2004psychology, tulving197212,delange2021continual}, is also important for artificial intelligence. 
In real-world scenarios, machine learning systems often have to gradually comprehend their goals instead of having them pre-set. For example, they may be required to grasp new topics and concepts from online media or identify unknown objects in dynamically changing environments. Nevertheless, deep neural networks (DNNs) mainly excel with close-world scenarios, where the class sets of training coincide with that of testing data.
Upon learning new knowledge, these systems may either fail or drastically decline performance on previously learned tasks, a known phenomenon termed catastrophic forgetting \cite{Goodfellow2014AnEI, McCloskey1989CatastrophicII,boschini2022class}. Hence, developing models that can continually adapt to new information represents a fundamental and challenging milestone for human-level intelligence. In this paper, we focus on challenging class-incremental learning (CIL).

To maintain the knowledge of old classes, previous research has primarily tackled catastrophic forgetting through three approaches: parameter regularization \cite{kirkpatrick2017overcoming,aljundi2018memory,li2017learning,dhar2019learning}, exemplar-based \cite{belouadah2018deesil,belouadah2019il2m,rebuffi2017icarl,castro2018end}, and architecture-based methods \cite{rusu2016progressive, mallya2018packnet, serra2018overcoming, yoon2017lifelong}. Among them, most state-of-the-art (SOTA) methods \cite{rebuffi2017icarl,hou2019learning,douillard2020podnet} are exemplar-based, which store a portion of old data and replay them when learning new classes. However, data storage is not an ideal solution due to various reasons. First, retraining on old data is inefficient and hard to scale to large-scale incremental learning due to limited memory resources \cite{parisi2019continual}. Second, concerns about privacy and security might not allow data storage in some situations \cite{li2020federated, appari2010information}. Additionally, from a biological perspective, storing raw data for memory is not aligned with human behavior \cite{kitamura2017engrams, kumaran2016learning, parisi2019continual}. An alternative is to learn deep generative models to generate pseudo-examples of old classes \cite{Shin2017ContinualLW, Wu2018MemoryRG, KemkerK18}, but generative models are inefficient and also prone to catastrophic forgetting \cite{Wu2018MemoryRG}.

To effectively learn new classes, existing methods often need a lot of labeled data.
However, the data that emerged in the real world is tremendously fast-paced, making it almost impossible to rely on manual annotations.
Ensuring label accuracy remains a labor-intensive task, even with the assistance of automated annotation tools.
Furthermore, domain-specific problems often require annotations from experts with relevant knowledge, which significantly increases both the annotation cost and difficulty.
The heavy reliance on annotation information undermines the performance of existing methods when labeled samples are scarce, and despite some few-shot CIL (FSCIL) \cite{zhang2021few} can reduce labeling requirements, they have a substantial performance gap compared to CIL methods.

\begin{figure}[t]
  \centering
  \includegraphics[width=\columnwidth]{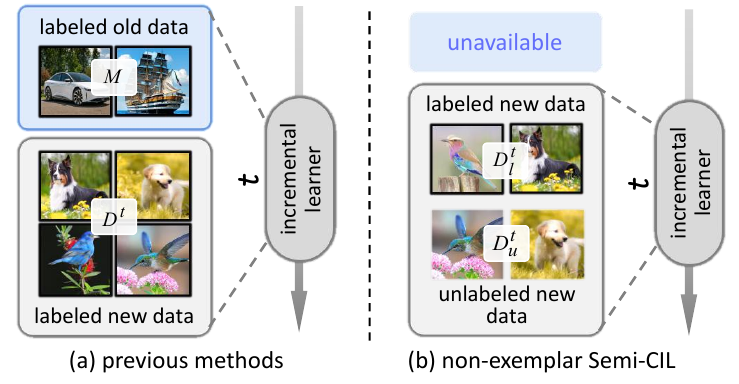}
  \vskip -0.05in
  \caption{Previous state-of-the-art CIL methods store and replay part of old data to alleviate the forgetting of previous knowledge, and require fully labeled new data to acquire new knowledge at each incremental stage $t$. In this paper, we consider a more challenging and practical non-exemplar semi-supervised CIL scenario where the old data is unavailable and only a few new examples are labeled (other unlabeled new data is accessible).}
  \vskip -0.05in
  \label{fig:1}
\end{figure}

Based on the above analysis, a natural question arises: \textbf{\textit{Without saving old samples, is it possible to attain superior CIL performance using just minimal labeled and abundant unlabeled data?}}
We define this as the non-exemplar semi-supervised CIL problem (as shown in Fig.~\ref{fig:1}), whose challenge lies in two aspects:
First, when learning new classes, the old data is not accessible. Consequentially, the network parameters will be updated based on new data, typically resulting in a substantial drop in the performance of old classes.
Second, the learning difficulty of new classes is heightened under the scarcity of labeled data, placing the model in a dilemma between recognizing new classes and maintaining the knowledge of old classes. 
To address this predicament, we break it down into the balance of feature extractor and classifier capabilities in recognizing new and old classes, proposing a simple yet effective two-stage method with contrastive learning and semi-supervised incremental prototype classifier (Semi-IPC).

For representation learning, recent methods use knowledge distillation (KD), ensuring new and old encoders remain similar in parameters and output.
However, old features might perform poorly on new classes, and overly learning new features can destabilize old class recognition.
These methods can only strike a middle ground through KD because features learned by supervised learning are highly category-specific. In light of this, we propose to leverage contrastive learning to
learn rich and task-agnostic feature representations, creating a substantial feature space for both old and new classes.
To learn a balanced classifier, we propose Semi-IPC which incrementally learns a prototype classifier from partially labeled new data while maintaining the knowledge of old classes. The prototype classifier has the superiority of indicating open-world knowledge \cite{yang2020convolutional}, leading to large separability between old and new classes. To further reduce forgetting, we leverage prototype resampling \cite{zhu2021prototype} to generate and replay pseudo-old samples in the feature space. Last but not least, unsupervised regularization helps the model learn new classes from unlabeled data. 
By balancing both features and classifiers, our method effectively addresses the non-exemplar semi-supervised CIL challenge, overcoming the limitations of previous methods that require storing old data and fully labeled new data. 

Our primary contributions can be summarized as follows:
\begin{enumerate}
\item[$\bullet$] We present the challenging but practical non-exemplar semi-supervised class-incremental learning setting to expand and prompt the CIL research.
\item[$\bullet$] We propose a simple and effective two-stage framework for the non-exemplar Semi-CIL challenge with contrastive learning and semi-supervised incremental prototype learning.
\item[$\bullet$] Extensive experimental results demonstrate that the proposed non-exemplar Semi-CIL method outperforms many CIL, FSCIL, and Semi-CIL methods, requiring less than 1\% labeled data and eliminating the need of storing old data.
\end{enumerate}

The rest of this paper is organized as follows: Related work is overviewed in Section \ref{sec: 2}. The Semi-CIL problem is defined in Section \ref{sec: 3}. Our proposed method is detailed in Section \ref{sec: 4}. Section \ref{sec: 5} presents the experimental results, and concluding remarks are given in Section \ref{sec: 6}.

\section{Related Work}
\label{sec: 2}

\subsection{Class Incremental Learning}
Existing CIL studies mainly learn new classes while retaining the knowledge of old class based on parameters regularization, data replay, and architecture expansion.

\textbf{\textit{Parameter-based methods}} can be divided into explicit \cite{kirkpatrick2017overcoming,zenke2017continual,aljundi2018memory} and implicit \cite{li2017learning,dhar2019learning,liu2020more} methods. Explicit methods limit the impact of new learning on key parameters of the model, while implicit methods maintain the consistency of model outputs through knowledge distillation \cite{hinton2015distilling}.
\textbf{\textit{Data replay-based methods}}, also known as exemplar-based methods, alleviate forgetting by saving and jointly training with a portion of old class data. This includes methods using old data directly \cite{belouadah2018deesil,belouadah2019il2m} and those introducing distillation loss to preserve old knowledge \cite{rebuffi2017icarl,castro2018end,douillard2020podnet,hou2019learning}, as well as strategies using generative models to synthesize raw data \cite{kamra2017deep,shin2017continual} or deep feature instances \cite{zhu2023} of old class.
\textbf{\textit{Architecture-based methods}} dynamically extend the network structure in incremental learning \cite{yang2022dynamic, mallya2018packnet, serra2018overcoming, yoon2017lifelong}. They prevent forgetting of old tasks by freezing original parameters and allocating new branches for new tasks. For example, DER \cite{yan2021dynamically}, FOSTER \cite{wang2022foster}, Dytox \cite{douillard2022dytox} and DNE \cite{hu2023dense} dynamically expand feature extractors to retain old knowledge when learning new classes, showing strong performance in CIL.

Recent works such as L2P \cite{wang2022learning}, DualPrompt \cite{wang2022dualprompt}, CODA-Prompt \cite{smith2022coda}, and S-iPrompts \cite{wang2022s} employ pre-trained models and learn a set of small, adjustable prompts for each task. However, directly leveraging the open-sourced pre-trained model would lead to information leakage, which significantly impacts the performance of these methods \cite{kimtheoretical}.

\begin{figure*}[t]
  \centering
  \includegraphics[width=1.0\linewidth]{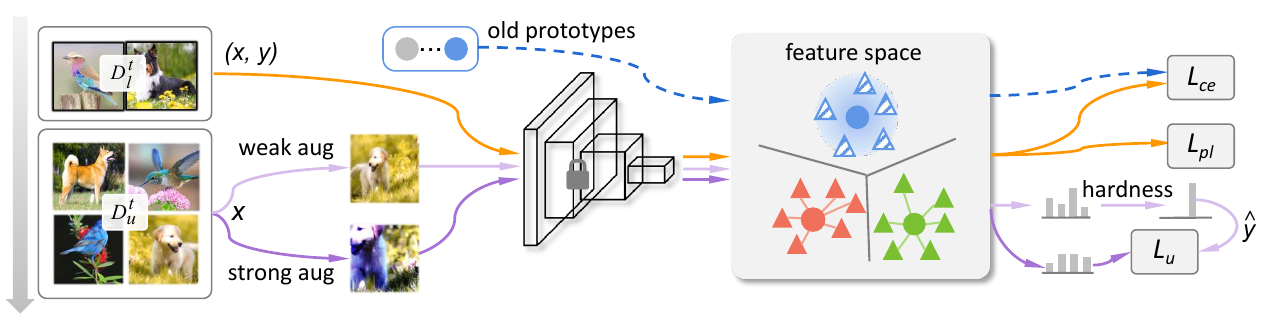}
  \vskip -0.1in
  \caption{Illustration of our two-stage framework for Semi-CIL. The feature extractor is learned via contrastive learning and then frozen in incremental learning process. A unified prototype classifier is updated incrementally with both supervised and unsupervised loss.
  Our method is non-exemplar based, simple but effective.}
  \label{fig:2}
\end{figure*}
\subsection{Few-shot Class-incremental Learning}
Few-shot Class-incremental Learning (FSCIL) \cite{tao2020few} aims to reduce the annotation cost of new classes. Initially, it requires all annotation for $N_0$ categories, but in the incremental learning phase, only $K$  labeled samples are needed for $N$ new categories, termed $N$ way $K$ shot. TOPIC  \cite{tao2020few} introduces a topology-preserving knowledge increment framework using a neural gas network, effectively integrating the preservation of old knowledge with the learning of new knowledge.  Recent works in FSCIL include methods based on structure and knowledge distillation. Among them, CEC \cite{zhang2021few} employs a graph model to propagate context among classifiers, adopting an effective strategy for decoupled learning of representations. SPPR \cite{zhu2021self} updates prototypes through a relationship matrix to enhance the extensibility of feature representation, and ERL \cite{dong2021few} proposes an Example Relation Distillation Incremental Learning framework further balances the stability and plasticity of the model during incremental learning process. 

Besides standard FSCIL, there are some studies that introduce unlabeled samples of new classes and gradually incorporate them and their pseudo-labels into the training set, significantly improving model performance with data quantities far exceeding traditional $K$-shot per class. This setting is known as Semi-Supervised FSCIL (Semi-FSCIL), with representative methods like Us-KD \cite{cui2022uncertainty} and UaD-CE \cite{cui2023uncertainty} optimizing the knowledge distillation process through unlabeled samples and their uncertainty. Although these methods have made significant progress beyond FSCIL, they face two major challenges: \textbf{(1)} they require full labeled data in the initial phase, leading to approximately half of the categories needing complete annotation; \textbf{(2)} there is still a significant gap compared to recent CIL methods. 

In this paper, we achieve the best performance in CIL and FSCIL settings through contrastive learning and semi-supervised incremental prototype classifiers, significantly outperforming existing methods.

\subsection{Contrastive Learning}
Self-supervised learning is widely used for feature extraction from unlabeled data. The primary technique, contrastive learning, generates positive and negative sample pairs through image enhancement, training models to produce similar representations for positive samples and distinct ones for negative samples, such as CPC \cite{oord2018representation}, SimCLR \cite{chen2020simple}, and MoCo \cite{he2020momentum}. 
\textbf{\textit{Negative-free methods}}, like BYOL \cite{grill2020bootstrap} and SimSiam \cite{chen2021exploring}, only use positive pairs to learn representations, thus avoiding the need for negative samples. Furthermore, recent studies included \textbf{\textit{Cluster-based methods}} (e.g., SwAV \cite{caron2020unsupervised} and DINO \cite{caron2021emerging}) and \textbf{\textit{Redundancy-based methods}} (e.g., BarlowTwins \cite{zbontar2021barlow} and VicReg \cite{bardes2022vicreg}). Specifically, cluster-based approaches use unsupervised clustering algorithms to align representations with prototypes, while redundancy reduction methods decrease the correlation between different representations, building on SimCLR \cite{chen2020simple}. For example, BarlowTwins optimizes the cross-correlation matrix to an identity matrix to align positive samples and orthogonalize negative ones, while VicReg combines variance, invariance, and covariance regularization in a unified framework.


\section{Preliminaries}
\label{sec: 3}
In a Semi-CIL scenario, tasks with non-overlapping class sets are learned sequentially, aiming for a unified model for all seen classes. Formally, at each incremental task $t$, dataset $\mathcal{D}^{t}$ is provided, consisting of labeled subset $\mathcal{D}^{t}_{l} = \{(\bm{x}_{i}^{t}, y_{i}^{t})\}^{n_l^t}_{i=1}$ and unlabeled subset $\mathcal{D}^{t}_{u} = \{\bm{x}_{i}^{t}\}^{n_u^t}_{i=1}$, where $\bm{x} \in \mathcal{X}$ is an input sample and $y \in \mathcal{C}_{t}$ is its label. $\mathcal{C}_{t}$ represents the class set at task $t$, with $\mathcal{C}_{i} \bigcap \mathcal{C}_{j} = \emptyset$ for $i \neq j$. Notably, we only have a few labeled data, i.e., $|\mathcal{D}^{t}_u|>>|\mathcal{D}^{t}_l|$. 
We denote DNN-based model with two components: \textbf{(1)} The feature extractor $f_{\bm{\theta}}: \mathcal{X} \rightarrow \mathcal{Z}$, with parameters $\bm{\theta}$, transforms input $\bm{x}$ into a feature vector $\bm{z} = f_{\bm{\theta}}(\bm{x}) \in \mathbb{R}^{d}$ in a deep feature space $\mathcal{Z}$. \textbf{(2)} The classifier $g_{\bm{\varphi}}: \mathcal{Z} \rightarrow \mathbb{R}^{|\mathcal{C}_{1:t}|}$, with parameters $\bm{\varphi}$, outputs a probability distribution $g_{\bm{\varphi}}(\bm{z})$ as the prediction for $\bm{x}$.
During incremental stage $t$, the goal is to minimize the loss $\ell$ on dataset $\mathcal{D}^{t}$, ensuring no negative impact on, and potentially improving previously acquired knowledge \cite{Aljundi2019ContinualLI}.

As analyzed in Sec. \ref{sec:introduction}, the difficulties of Semi-CIL lie in two folds. On the one hand, the old class data is unavailable, making it challenging for the model to retain old knowledge and prone to forgetting. On the other hand, there are only a few labeled data of new classes, making it hard to learn new knowledge.
In conclusion, the unavailability of data and insufficient labeling cause problems in incremental learning, including poor recognition of new classes and severe forgetting of old knowledge.
Therefore, balanced recognition of both new and old classes in $f_{\bm{\theta}}$ and $g_{\bm{\varphi}}$ becomes even more important.

\section{Our Approach}
\label{sec: 4}
\subsection{Overview of the framework}
As illustrated in Fig. \ref{fig:2}, the proposed two-stage Semi-CIL framework comprises a fixed contrastive learning feature extractor and a semi-supervised incremental prototype classifier (Semi-IPC). In the first stage, we extract rich, task-agnostic feature representations through contrastive learning, creating a shared feature space for all categories. This is compatible with a wide array of contrastive learning techniques \cite{grill2020bootstrap, he2020momentum, chen2020simple, caron2020unsupervised, zbontar2021barlow, chen2021exploring}. In the second stage, Semi-IPC continuously learns and updates prototypes for each category. Classification during inference is achieved by computing the Euclidean distance between samples and their nearest prototypes. The novelty stems from two aspects: first, different from existing CIL, Semi-CIL, and FSCIL methods that typically rely on knowledge distillation for feature learning, our feature extractor learns category-agnostic representations through contrastive learning and is subsequently frozen. Second, Semi-IPC learns and updates prototypes with limited labeled data and a large number of unlabeled data, a departure from nearest-mean-of-examples (NME) or cosine linear classifiers. Moreover, our method is example-agnostic, requiring less than 1\% labeled data, making it simple and effective.

\subsection{Contrastive Representations Learning}
\label{sec: 4.1}
A primary difficulty in Semi-CIL is managing feature representations that suit both new and old classes, ensuring adaptability to new classes without diminishing the performance on old classes.
We suggest that this paradox can be reconciled through contrastive learning, as it is capable of generating features that apply to both new and old classes. 
Take BYOL \cite{grill2020bootstrap} as an example, its learning process can be summarized as follows.
BYOL includes an online network \( f_{\theta} \) and a target network \( f_{\xi} \), parameterized by \( \theta \) and \( \xi \) respectively. The target network is the exponential moving average of the online network, following the update rule:
\begin{equation}
\label{eq1}
\xi \leftarrow m\xi + (1 - m)\theta, 
\end{equation}
where \( m \) is the momentum coefficient. Both networks comprise a backbone \( g \) and a projection head \( h \).
For an input image \( \bm{x} \), we generate two augmented views \( \bm{x}_i \) and \( \bm{x}_j \), with the goal of minimizing the distance between vectors \( \bm{z}_i \) and \( \bm{z}_j \). This process is expressed through:
\begin{equation}
\label{eq2}
\mathcal{L}(\theta) = \frac{1}{N} \sum_{i=1}^{N} \Big\lVert \frac{\bm{z}_i}{\lVert \bm{z}_i \rVert} - \frac{\bm{z}_j^{\xi}}{\lVert \bm{z}_j^{\xi} \rVert} \Big\rVert^{2},
\end{equation}
where \( N \) is the batch size, and the vectors \( \bm{z}_i \) and \( \bm{z}_j^{\xi} \) are defined as \( \bm{z}_i = h(f_{\theta}(\bm{x}_i)) \) and \( \bm{z}_j^{\xi} = h(f_{\xi}(\bm{x}_j)) \) respectively. 

Contrastive learning trains the model to generate consistent feature representations across various augmented views, enhancing both \textbf{\textit{feature richness}} and \textbf{\textit{generalization}} capabilities across classes. These enhancements are essential for fulfilling the requirements of continual learning. We quantify these aspects using intrinsic dimensionality (PC-ID) and feature space uniformity (FSU).

\subsubsection{Feature Richness}
In contrastive learning, feature extraction extends past predefined classes, and the richness of these features is assessed by PC-ID, the minimal coordinate count necessary for describing data without significant information loss 
\cite{ansuini2019intrinsic,yu2020learning,chen2021understanding}. Technically, the estimation of this dimensionality within the feature space is performed by PCA analysis \cite{pearson1901liii} of the normalized feature covariance matrix, $\Sigma$. With the eigenvalues $\lambda_i$ of $\Sigma$ in descending order, the cumulative normalized eigenvalue sum $P(k)$ is expressed as:
\begin{equation}
\label{eq4}
P(k) = \frac{\sum_{i=1}^{k} \lambda_i}{\sum_{i=1}^{N} \lambda_i},
\end{equation}
where $N$ is the eigenvalue count. The PC-ID is:
\begin{equation}
\text{PC-ID} = \min k \in \mathbb{N} : P(k) \geq 0.9 .
\end{equation}

\begin{figure}[t]
	\begin{center}
		\centerline{\includegraphics[width=0.95\columnwidth]{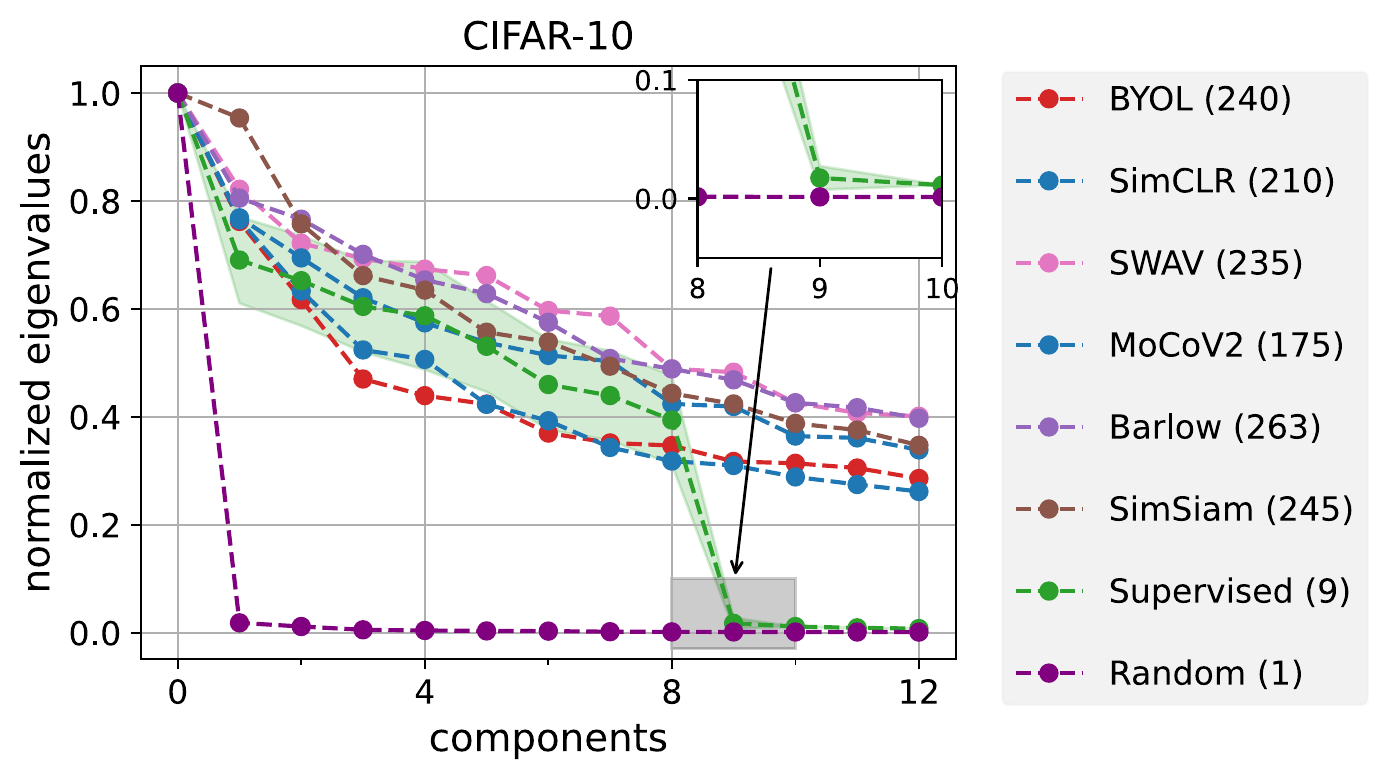}}
		\vskip -0.07in
		\caption{The normalized singular values of the feature matrix show descending curves of the top 12 eigenvalues. In supervised learning, they drop to 0 at $N-1$. SSL has a higher PC-ID with a slower curve decline.}
		\label{figure-3}
	\end{center}
	\vskip -0.25in
\end{figure}

We compare common supervised learning and contrastive learning on CIFAR-10 \cite{krizhevsky2009learning} dataset. We trained ResNet18 \cite{he2016deep} network using 6 contrastive learning methods (BYOL \cite{grill2020bootstrap}, MoCoV2 \cite{he2020momentum}, SimCLR \cite{chen2020simple}, SwAV \cite{caron2020unsupervised}, BarlowTwins \cite{zbontar2021barlow}, and SimSiam \cite{chen2021exploring}) and analyzed PC-ID of feature space. As shown in Fig. \ref{figure-3}, supervised learning focuses on features for separating current classes, resulting in PC-ID $=K-1$ \cite{chen2021understanding}, where $K$ is the number of classes. In contrast, the normalized eigenvalue curve of contrastive learning declines more gradually, indicating that it learns a feature space with a higher intrinsic dimension.

\begin{figure}[h]
	\begin{center}
            \centerline{\includegraphics[width=0.95\columnwidth]{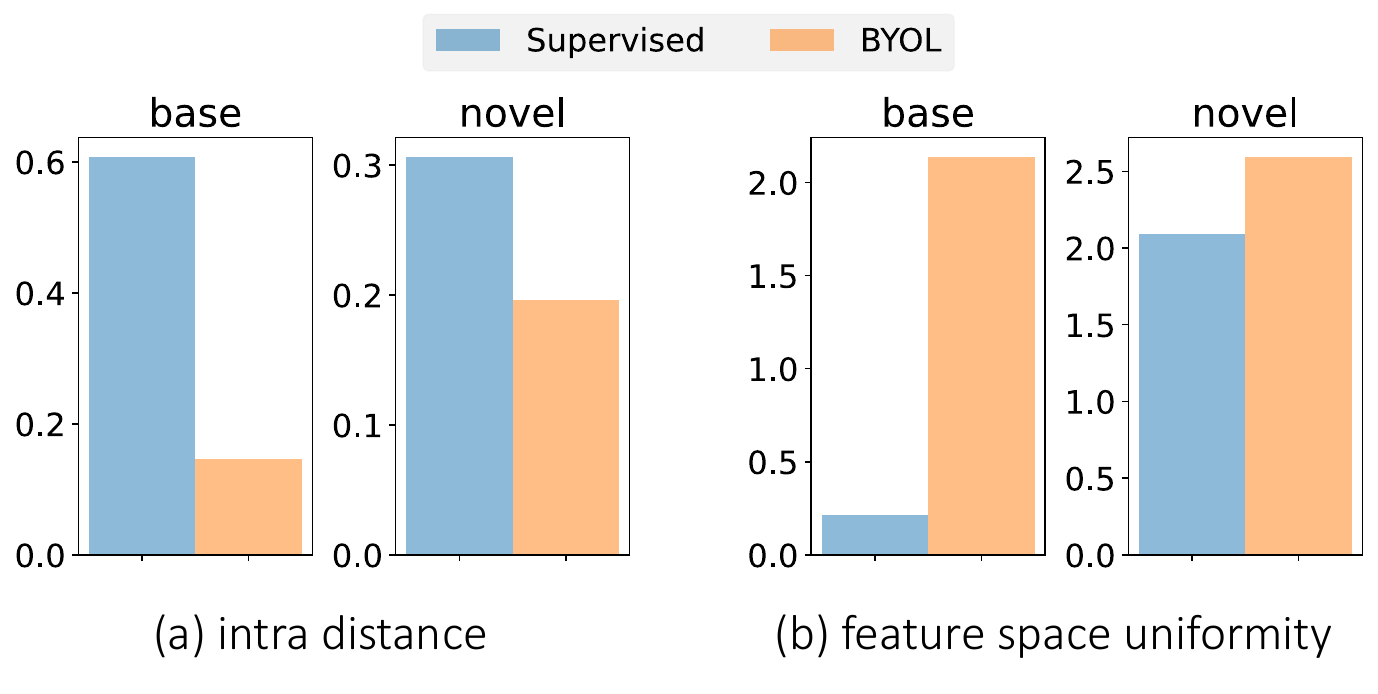}}
		\vskip -0.07in
		\caption{Contrastive learning results in smaller intra-distance on novel classes, and high feature space uniformity.}
		\label{figure-5}
	\end{center}
	\vskip -0.25in
\end{figure}
\subsubsection{Novel Class Generalization}

We assess class distributions quantitatively by calculating the average inter-class ($\pi_{inter}$) and intra-class ($\pi_{intra}$) distances, defined as follows:
\begin{equation}
\label{eq5}
\pi_{inter} =  \frac{1}{Z_{inter}} \sum_{y_{l},y_{k},l \neq k}d(\mu(\bm{z}_{y_{l}}), \mu(\bm{z}_{y_k})),
\end{equation}
\begin{equation}
\label{eq6}
\pi_{intra} =  \frac{1}{Z_{intra}} \sum_{y_{l} \in y} \sum_{\bm{z}_{i},\bm{z}_{j} \in \bm{z}_{y_{l}},i \neq j}d(\bm{z}_{i}, \bm{z}_{j}).
\end{equation}
where $\bm{z}_{y_{l}}=\{\bm{z}_{i}:=f_{\theta}(\bm{x}_{i})|\bm{x}_{i} \in \bm{x}, y_{i}=y_{l}\}$ denotes the sample embeddings for class $y_{l}$, $d$ is the cosine distance, $\mu(\bm{z}_{y_{l}})$ is the mean embedding, and $Z_{intra}$ and $Z_{inter}$ are normalization constants.

Fig.~\ref{figure-5} demonstrates the effectiveness of contrastive learning in reducing $\pi_{intra}$ for unknown classes, indicating its superior generalization. Supervised learning shows a more compact feature distribution for trained classes, which may hinder the generalization of representations to new classes. Features insignificant to the current task might be critical for future tasks. Therefore, learning overly compressed representations could overlook key information beneficial for future tasks.
Considering this, we introduce the feature space uniformity (FSU) metric $\pi_{ratio} = \pi_{intra} /\pi_{inter}$ \cite{roth2020revisiting}. As shown in Fig.~\ref{figure-5} (b), a higher FSU value indicates the potentiality of contrastive learning for generalization to new classes, aligning with the conclusions in \cite{roth2020revisiting}.

\subsection{Semi-Supervised Incremental Prototype Classifier}
Another issue in Semi-CIL is ensuring the classifier retains old information while integrating new information effectively, especially when sample labeling is inadequate. It becomes quite challenging to maintain high recognition performance while achieving balanced classification of new and old classes.
%
Therefore, maintaining a balance in recognizing new and old classes is essential, which sets semi-CIL apart from traditional Semi-supervised learning methods.
Our solution is the Semi-Supervised Incremental Prototype Classifier (Semi-IPC), composed of the \textbf{\textit{Incremental Prototype Classifier}}, \textbf{\textit{Prototype Resampling}}, and \textbf{\textit{Prototype Unsupervised Regularization}}. The following sections describe each component and how they work together.

\subsubsection{Incremental Prototype Classifier}
\label{sec: 4.2.1}
Semi-IPC uses the Incremental Prototype Classifier (IPC) to integrate new data while retaining recognition of old classes. In our model, prototypes are represented as $ \varphi=\{\bm{\varphi}_i | i=1,2,...,|\mathcal{C}_{1:t}|\} $, existing as vectors in feature space $ \mathbb{R}^d $. An input $ \bm{x} $ is classified by extracting its features $ f_{\theta}(\bm{x}) $ and comparing the Euclidean distance to each prototype, calculated as $ d_i(\bm{x}) = ||f_{\theta}(\bm{x})-\bm{\varphi}_i||_2^2 $.

\textbf{\textit{Discriminative loss and Regularization loss.}}
In the feature space, the probability that a sample $(\bm{x},y)$ matches the prototype $\bm{\varphi}_i$ of class $i$ is calculated by a distance-based softmax, formulated as:
\begin{equation}
\label{eq7}
p(\arg\min_{i=1}^{|\mathcal{C}_{1:t}|} d_i(\bm{x}) = y)=\frac{e^{-\gamma d_i(\bm{x})}}{\sum_{k=1}^C e^{-\gamma d_k(\bm{x})}},
\end{equation}
Where $\gamma$ is a temperature scalar that affects the sparseness of the class distribution. Based on this probability, we use cross-entropy as the discriminative loss:
\begin{equation}
  \label{eq8}
  \mathcal{L}_{ce}\big((\bm{x},y);\varphi\big) = -\log p(\arg\min_{i=1}^{|\mathcal{C}_{1:t}|} d_i(\bm{x}) = y).
\end{equation}
To prevent over-adjustment of prototypes leading to deviation from the sample distribution, we introduce Prototype Learning Loss (PL). PL acts like a generative model to enhance the classifier's generalization capability in incremental learning. PL aims to bring prototypes and corresponding samples closer, reducing confusion with newly added classes. It is defined as:
\begin{equation}
\label{eq9}
\mathcal{L}_{pl}\big((\bm{x},y);\varphi\big)=||f_{\theta}(\bm{x})-\bm{\varphi}_y||_2^2.
\end{equation}
Combining both discriminative and regularization losses, the loss function for IPC is defined as:
\begin{equation}
\label{eq10}
  \mathcal{L}_{ipc} = \mathcal{L}_{ce}+\lambda \cdot \mathcal{L}_{pl}.
\end{equation}
where $\lambda$ is a hyper-parameter.

\textbf{\textit{Incremental updating.}} In Semi-CIL, to mitigate the adverse impact of adding new prototypes on the decision boundary, we design an incremental update strategy. Specifically, when new prototypes of classes $\varphi_{new}$ are added, the existing prototypes, $\varphi_{old}$, contribute to forward propagation for loss computation but are excluded from backpropagation. It can be described by the calculation of distance $d_i$ as follows:
\begin{equation}
\label{eq11}
\begin{aligned}
  \varphi &= [\textbf{\textit{StopGrad}}(\varphi_{old}), \varphi_{new}], \\
  d_i(x) &= ||f_{\theta}(\bm{x})-\bm{\varphi}_{i}||_2^2.
\end{aligned}
\end{equation}
This strategy maintains the decision boundary between old and new classes and avoids altering the position of old prototypes, thus reducing the problem of forgetting.

\textbf{\textit{Inference.}} At test time, the given input $\bm{x}$ is classified by finding the prototype with the smallest Euclidean distance:
\begin{equation}
\label{eq12}
  \bm{x} \in \arg\min_{i=1}^{|\mathcal{C}_{1:t}|} d_i(\bm{x}),
\end{equation}
where $\mathcal{C}_{1:t}$ is the set of learned classes.

\begin{figure}[t]
	\begin{center}
		\centerline{\includegraphics[width=0.95\columnwidth]{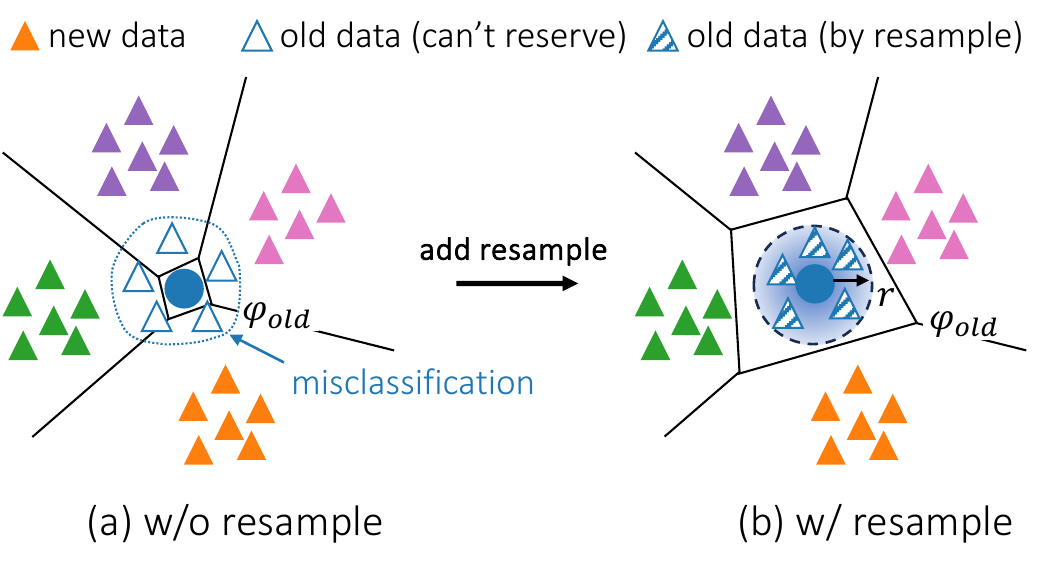}}
		\vskip -0.07in
		\caption{Without saving old data, prototype resample generates pseudo-feature instances around old class prototypes, maintaining decision boundaries between old and new classes.}
		\label{figure-6}
	\end{center}
	\vskip -0.15in
\end{figure}

\subsubsection{Prototype Resample}
As demonstrated in Fig. \ref{figure-6}, non-exemplar Semi-CIL uses old prototypes, a small set of new class data, and unlabeled data for classifier training. 
The absence of old samples in joint learning inevitably impacts the classification decisions between new and old classes. Thanks to the prototype classifier, the model can generate pseudo-feature instances of old classes by prototype resampling, i.e., for each old class $k \in \mathcal{C}_{old}$,  pseudo-feature instance $\widetilde{\bm{z}}_{k}$ can be generated from prototype $\bm{\varphi}_{k} $ as follows:
\begin{equation}
\label{eq13}
\begin{aligned}
\widetilde{\bm{z}}_{k} = \bm{\varphi}_{k} + r \cdot \bm{e},
\end{aligned}
\end{equation}
Where $r$ is a scale factor controlling the uncertainty of resampling, and $\bm{e} \sim \mathcal{N}(\bm{0}, 1)$ is a Gaussian noise vector with the same dimension as the prototype.
We use the average feature variance in the initial task to estimate the value of $r$:
\begin{equation}
\label{eq14}
\begin{aligned}
r^2 = \frac{1}{|\mathcal{C}_{t=1}| \cdot d}\sum\nolimits_{k=1}^{|\mathcal{C}_{t=1}|}\textit{tr}(\bm{\Sigma}_{k}). 
\end{aligned}
\end{equation}
After that, Eq. (\ref{eq10}) is applied for jointly training on new feature instances $f_{\theta}(x)$ and old pseudo-feature instances $\widetilde{\bm{z}}$ within the Semi-IPC framework,
supporting joint classification among base and novel classes.

\subsubsection{Prototype Unsupervised Regularization}
In real-world incremental learning scenarios, the amount of labeled data is typically limited, while unlabeled data is easy to acquire. Although unsupervised learning methods have been extensively studied, they are rarely explored in incremental learning. In this paper, we introduce prototype unsupervised regularization for Semi-IPC and demonstrate the significant performance improvements.

\textbf{\textit{Strong and weak augmentation.}} Data augmentation is a key part of unsupervised learning, aiming to train models for consistent predictions across various transformations of samples.
Our method adopts a dual strategy of strong and weak augmentation, a concept derived from semi-supervised works \cite{xie2020unsupervised, berthelot2019remixmatch, sohn2020fixmatch}.
Specifically, weak augmentation $\mathcal{A}_{w}(x)$ includes basic image operations such as cropping, flipping, or displacement. In contrast, strong augmentation $\mathcal{A}_{s}(x)$ involves more aggressive actions like significant distortion, noise addition, or complex filter.
We perform strong augmentation based on RandAugment \cite{cubuk2020randaugment}, where each sample's type and extent of transformations are randomly chosen.

\begin{algorithm}[t]
\caption{Semi-IPC Learning Algorithm}
\label{alg1}
	$\bm{\Theta}^{0} = \{ \bm{\theta}, \bm{\varphi}^{0}\} $\;
	$\bm{\theta} \leftarrow$ contrastive pre-trained feature extractor\;
	$\bm{\varphi}^{0}\ = \emptyset $, $r = 0$\;
\ForEach{incremental stage $t$}{
\KwIn{prototypes $\varphi^{t-1}$, new data $\mathcal{D}^{t} = \mathcal{D}_u^t \cup \mathcal{D}_l^t$;}
\KwOut{prototypes $\varphi^{t}$;}

\textcolor{gray}{\# prototypes initial}

$\bm{\varphi}^{new} \leftarrow$ initialized by class means\;
$\bm{\varphi}^{t} \leftarrow [\textbf{\textit{StopGrad}}(\bm{\varphi}^{t-1}),\bm{\varphi}^{new}]$\;
 
\eIf{$t = 1$}{
$r \leftarrow$ compute radial scale using Eq.~(\ref{eq14})\;
$\mathcal{Z}_{old}^{t} \leftarrow \emptyset $ \;
}{
\textcolor{gray}{\# prototype resample for old classes}

$\mathcal{Z}_{old}^{t}  \leftarrow$ $\{\widetilde{\bm{z}}_{t-1}, \mathcal{C}_{t-1} \ |\  \widetilde{\bm{z}}_{t-1} = \bm{\varphi}^{t-1} + r \cdot \bm{e}\}$ \;
}
\For{$(\bm{z}_k,y_k)$ in $\{(f_{\theta}(\bm{x}),y)| (\bm{x},y) \in \mathcal{D}_l^t\} \cup \mathcal{D}_{old}^t$}{
$\mathcal{L}_{ipc} \leftarrow$ compute supervised loss by Eq.~(\ref{eq10}) \;
}
\For{$x_k$ in $\mathcal{D}_{u}^t$} {
$\mathcal{L}_{u} \leftarrow$ compute unsupervised loss by Eq.~(\ref{eq18}) \;
}
train $\bm{\varphi}^{t}$ by minimizing $\mathcal{L}_{ipc}+\mathcal{L}_{u}$ ins Eq.~(\ref{eq19})\;
}
\end{algorithm}
\textbf{\textit{Prototype unsupervised regularization.}} At each iteration, we use maximum softmax probability to select reliable, confident unlabeled samples $\mathcal{U}$ in each minibatch as follows:
\begin{equation} 
\label{eq16}
\mathcal{U} = \{\bm{x} | \max_i p(\mathcal{A}_{w}(\bm{x})|\bm{x}) > \tau\}, 
\end{equation}
Where $\tau$ is the threshold for select sample $\bm{x}$ in unlabeled minibatch if it's close enough to a class prototype. Then, the pseudo-label of $\bm{x}$, denoted as $\hat{y}$, is computed based on the nearest prototype, calculated by:
\begin{equation} 
\label{eq17}
\hat{y} = \arg\min_{i=1}^{|\mathcal{C}_{1:t}|} d_i(\mathcal{A}_{w}(\bm{x})). 
\end{equation}
The $\hat{y}$ is then assigned as a pseudo-label of the strong augmentation sample $\mathcal{A}_{s}(\bm{x})$.
By combining with strong transformations $\mathcal{A}_{s}(\bm{x})$, we generate a training batch of unlabelled samples and their pseudo-labels $\{(\mathcal{A}_{s}(\bm{x}), \hat{y}),\bm{x} \in \mathcal{U}\}$.
After obtaining the unsupervised training set $\mathcal{U}$, Semi-IPC pushes $\mathcal{A}_{s}(\bm{x})$ to the corresponding pseudo-label $\hat{y}$ by optimizing the cross-entropy loss $\mathcal{L}_{ce}$ as follows:
\begin{equation}
\label{eq18} 
\begin{aligned} 
\mathcal{L}_{u}(\bm{x};\varphi) &= \mathcal{L}_{ce}\big((\mathcal{A}(\bm{x}),\hat{y});\varphi\big).
\end{aligned}
\end{equation}

\subsection{Integrated Learning Objective} \label{sec:objective}
Contrastive learning yielded rich features with high generalization capabilities. Combined with Semi-IPC, the model can learn new classes from a few labeled and many unlabeled samples, preventing catastrophic forgetting. The integration of contrastive learning and Semi-IPC optimizes the balance in feature extraction and classification across new and existing classes. The learning objective at each incremental stage is: 
\begin{equation} 
\label{eq19} 
\begin{aligned} 
\mathcal{L} = \mathcal{L}_{\text{ipc}} +  \mathcal{L}_{u} 
= \mathcal{L}_{ce} + \lambda \cdot \mathcal{L}_{pl} + \mathcal{L}_{u}
\end{aligned}
\end{equation}
The formula includes a hyperparameter $\lambda$ . $\mathcal{L}_{\text{ipc}}$ and $\mathcal{L}_{u}$ are detailed in Eq. (\ref{eq10}) and Eq. (\ref{eq18}), respectively.
The overall learning process is summarized in Algorithm \ref{alg1}.

\section{Experiments}
\label{sec: 5}
\subsection{Setup}
\noindent
\textbf{Dataset.}
We conducted experiments on 4 benchmark datasets including ImageNet-100 \cite{deng2009imagenet}, CIFAR-100 \cite{krizhevsky2009learning}, miniImageNet \cite{vinyals2016matching} and CUB-200 \cite{AGEM}. 
Among them, ImageNet-100 and CIFAR-100 are widely used in CIL literature, while miniImageNet and CUB-200 are commonly used in FSCIL. 
We employ two schemes for dividing datasets: a uniform division into $\bm{T}$ tasks (i.e., $\bm{B=0}$), following the setting of \cite{yan2021dynamically, ahn2021ss, wang2022foster}, or using a majority portion for the base task (e.g., $\bm{B=50, 60, 100}$) with the rest of the classes divided evenly into $\bm{T}$ tasks \cite{Rebuffi2017iCaRLIC, Douillard2020SmallTaskIL, cui2023uncertainty}. In exemplar-based methods, the number of retained samples/class is represented as $\bm{R}$.
Table \ref{table-1} shows the number of labeled ($\bm{|\mathcal{D}_{s}|}$) and unlabeled samples ($\bm{|\mathcal{D}_{u}|}$) for each category in these datasets for CIL, Semi-CIL, and FSCIL methods.

\begin{table}[t]
	\caption{Dataset setup in CIL, Semi-CIL, and FSCIL. Task 1 (base) of FSCIL utilizes all available data, while for novel classes, only 5 samples per class are used.}
	\vskip -0.13in
	\label{table-1}
	\begin{center}
		\renewcommand\tabcolsep{4pt}
		\newcommand{\tabincell}[2]{\begin{tabular}{@{}#1@{}}#2\end{tabular}}
		\scalebox{1}{
			\renewcommand{\arraystretch}{1.1}
\begin{tabular}{lcccccc}
\toprule[1.3pt]
\multirow{2}{*}{\textbf{Dataset}} & \multicolumn{2}{c}{\textbf{CIL}} & \multicolumn{2}{c}{\textbf{Semi-CIL}} & \multicolumn{2}{c}{\textbf{FSCIL}} \\ \cmidrule(lr){2-3} \cmidrule(lr){4-5} \cmidrule(lr){6-7}
                                  & $|\mathcal{D}_{s}|$                & $|\mathcal{D}_{u}|$             & $|\mathcal{D}_{s}|$                & $|\mathcal{D}_{u}|$   & $|\mathcal{D}_{s}|$                & $|\mathcal{D}_{u}|$                  \\ \midrule
CIFAR-100 \cite{krizhevsky2009learning}                        & 500              & 0             & 5                & 495  & 5                & 0                \\
ImageNet-100 \cite{deng2009imagenet}                     & 1300             & 0             & 5                & 1295  & 5                & 0               \\
miniImageNet \cite{vinyals2016matching}                     & 600              & 0             & 5                & 595 & 5                & 0                       \\
CUB-200 \cite{AGEM}                          & -               & -                & 5               & 25 & 5               & 0  \\

\bottomrule[1.3pt]
\end{tabular}
		}
	\end{center}
	\vskip -0.1in
\end{table}
\begin{table*}[t]
    \caption{Comparisons of average and last incremental accuracies (\%) with the dataset evenly split into 5/10 tasks ($B=0, T=5/10$). CIL methods use all labeled data and BYOL pre-trained model.}
    \vskip -0.13in
    \label{table-2}
    \begin{center}
        \renewcommand\tabcolsep{4.5pt}
        \newcommand{\tabincell}[2]{\begin{tabular}{@{}#1@{}}#2\end{tabular}}
        \scalebox{1}{
            \renewcommand{\arraystretch}{1.37}
            \begin{tabular}{cclcccccccccccc}
                \toprule[1.3pt]
                & & & \multicolumn{4}{c}{\textbf{CIFAR-100}} & \multicolumn{4}{c}{\textbf{ImageNet-100}} & \multicolumn{4}{c}{\textbf{miniImageNet}} \\
                \cmidrule(lr){4-7} \cmidrule(lr){8-11} \cmidrule(lr){12-15}
                & & & \multicolumn{2}{c}{T=5} & \multicolumn{2}{c}{T=10} & \multicolumn{2}{c}{T=5} & \multicolumn{2}{c}{T=10} & \multicolumn{2}{c}{T=5} & \multicolumn{2}{c}{T=10} \\
                \cmidrule(lr){4-5}  \cmidrule(lr){6-7} \cmidrule(lr){8-9} \cmidrule(lr){10-11} \cmidrule(lr){12-13} \cmidrule(lr){14-15} 
                \multirow{-3}{*}{\textbf{$\bm{|D_{s}|}$}} & \multirow{-3}{*}{\textbf{Exemplar}} & \multirow{-3}{*}{\textbf{Method}} & Avg & Last & Avg & Last & Avg & Last & Avg & Last & Avg & Last & Avg & Last \\ 
                \midrule
                & & iCaRL \cite{Rebuffi2017iCaRLIC} & 69.13 & 53.82 & 62.17 & 46.23 & 69.73 & 53.72 & 58.56 & 43.06 & 53.88 & 35.88 & 48.58 & 29.86 \\
                & & UCIR \cite{Hou2019LearningAU} & 68.02 & 52.39 & 59.98 & 45.87 & 74.55 & 59.30 & 60.60 & 42.98 & 61.20 & 42.72 & 53.37 & 33.06 \\
                & & PODNet \cite{Douillard2020SmallTaskIL} & 70.34 & 53.52 & 61.63 & 46.08 & 75.22 & 60.90 & 66.28 & 50.64 & 62.50 & 44.36 & 53.95 & 33.22 \\
                & & FOSTER \cite{wang2022foster} & 65.44 & 48.75 & 64.88 & 54.24 & 66.70 & 48.84 & 67.91 & 58.36 & 60.95 & 46.36 & \underline{59.82} & \underline{47.28} \\
                & & DER \cite{yan2021dynamically} & \underline{75.69} & \underline{63.06} & \underline{66.66} & \underline{49.82} & \textbf{79.12} & \textbf{69.10} & \underline{73.81} & \underline{62.98} & \underline{66.26} & \underline{52.14} & 59.13 & 44.40 \\
                \multirow{-6}{*}{500\textasciitilde1300} & \multirow{-6}{*}{1000} & MEMO \cite{zhou2023model} & 67.05 & 55.52 & 59.48 & 49.59 & 72.44 & 63.46 & 65.07 & 53.66 & 57.54 & 42.64 & 46.05 & 32.70 \\
                \midrule
                \midrule
                & & LWF \cite{li2017learning} & 66.10 & 46.29 & 50.18 & 27.30 & 64.48 & 42.82 & 50.11 & 26.48 & 58.18 & 40.30 & 49.89 & 27.44 \\
                & & PASS \cite{Zhu_2021_CVPR} & 70.21 & 57.75 & 62.73 & 48.72
                                & 70.49 & 56.24 & 60.47 & 41.32 & 60.14 & 44.36 & 54.14 & 36.30 \\
                & & SSRE \cite{zhu2022self} & 63.67 & 48.75 & 60.06 & 46.29 & 62.86 & 47.34 & 52.32 & 35.28 & 58.19 & 44.06 & 50.30 & 33.34 \\
                \multirow{-4}{*}{500\textasciitilde1300} & \multirow{-4}{*}{0} & FeTrIL \cite{Petit_2023_WACV} & 65.62 & 50.45 & 62.98 & 49.05 & 64.99 & 49.12 & 53.34 & 34.06 & 52.41 & 36.56 & 49.70 & 31.82 \\
                \rowcolor{mygray}
                5 & 0 & Semi-IPC & \textbf{75.81} & \textbf{67.64} & \textbf{74.92} & \textbf{65.29} & \underline{76.00} & \underline{68.04} & \textbf{75.52} & \textbf{67.02} & \textbf{68.31} & \textbf{60.70} & \textbf{69.99} & \textbf{60.00} \\
                \bottomrule[1.3pt]
            \end{tabular}
        }
    \end{center}
\end{table*}

\begin{figure*}[t]
	\begin{center}
  		\centerline{\includegraphics[width=0.9\textwidth]{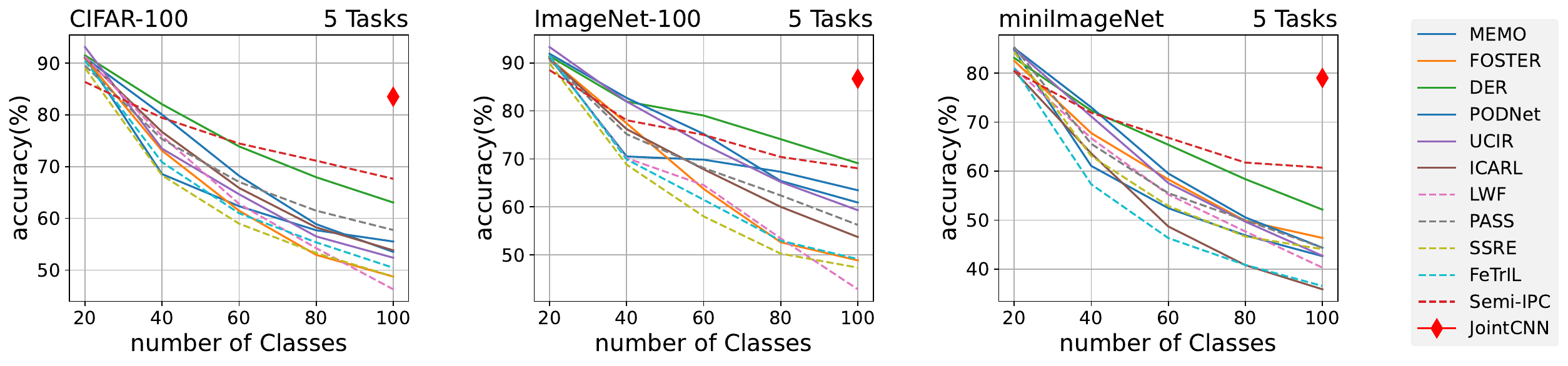}}
		\vskip -0.07in
		\caption{Results of classification accuracy on CIFAR-100, ImageNet-100 and miniImageNet which contains 5 sequential tasks. Dashed lines represent non-exemplar methods, and solid lines denote exemplar-based methods. Our non-exemplar method uses only 5 labeled samples per class ($|\mathcal{D}_{s}|=5$). }
		\label{figure-8}
	\end{center}
	\vskip -0.25in
\end{figure*}

\begin{figure*}[!t]
	\begin{center}
  		\centerline{\includegraphics[width=0.9\textwidth]{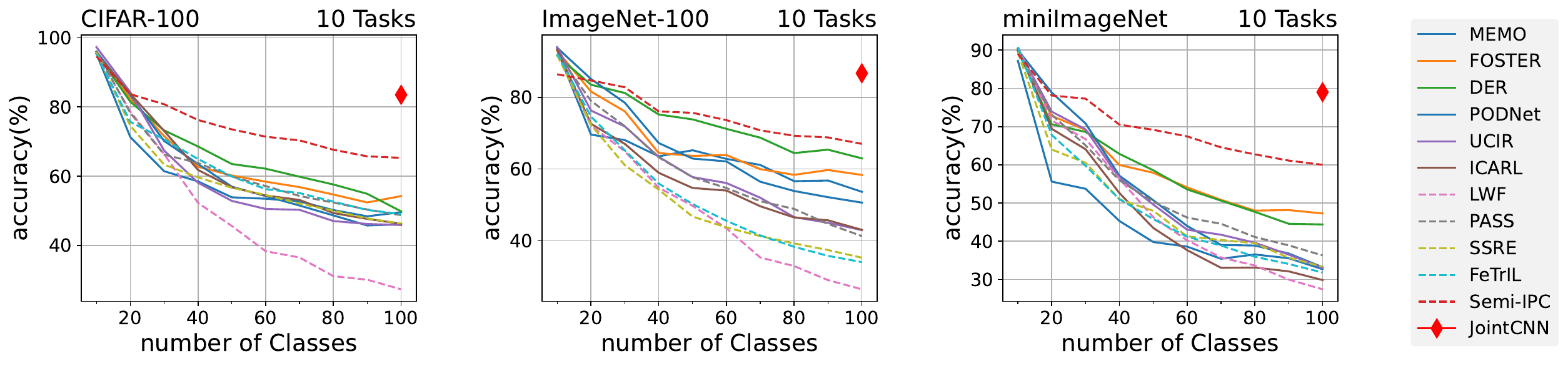}}
		\vskip -0.07in
		\caption{Results of classification accuracy on CIFAR-100, ImageNet-100 and miniImageNet which contains 10 sequential tasks.}
		\label{figure-9}
	\end{center}
	\vskip -0.1in
\end{figure*}

\vspace*{5pt}
\noindent
\textbf{Comparison methods.} We compare the proposed Semi-IPC with various baselines, including standard CIL, FSCIL, and Semi-CIL methods. 
\textbf{(1)} \textbf{\emph{CIL Methods:}} iCaRL \cite{Rebuffi2017iCaRLIC}, UCIR \cite{Hou2019LearningAU}, PODNet \cite{Douillard2020SmallTaskIL}, FOSTER \cite{wang2022foster}, DER (w/o pruning) \cite{yan2021dynamically}, MEMO \cite{zhou2023model},  LWF \cite{li2017learning}, PASS \cite{Zhu_2021_CVPR}, SSRE \cite{zhu2022self}, FeTrIL \cite{Petit_2023_WACV}, GeoDL \cite{simon2021learning}, Mnemonics \cite{liu2020mnemonics}, AANets \cite{liu2021adaptive}, CwD \cite{shi2022mimicking}, SSIL \cite{ahn2021ss}, DMIL \cite{tang2022learning}, AFC \cite{kang2022class}, DualNet \cite{pham2021dualnet},  BiMeCo \cite{nie2023bilateral}, EOPC \cite{wen2023optimizing}, Dytox \cite{douillard2022dytox} and DNE \cite{hu2023dense}.
\textbf{(2)} \textbf{\emph{FSCIL Methods:}} FACT \cite{zhou2022forward}, CEC \cite{zhang2021few}, SPPR \cite{zhu2021self}, IDLVQ-C \cite{chen2020incremental}.
\textbf{(3)} \textbf{\emph{Semi-FSCIL Methods:}} Us-KD \cite{cui2022uncertainty} and UaD-CE \cite{cui2023uncertainty}.

\begin{figure*}[t]
	\begin{center}
  		\centerline{\includegraphics[width=0.88\textwidth]{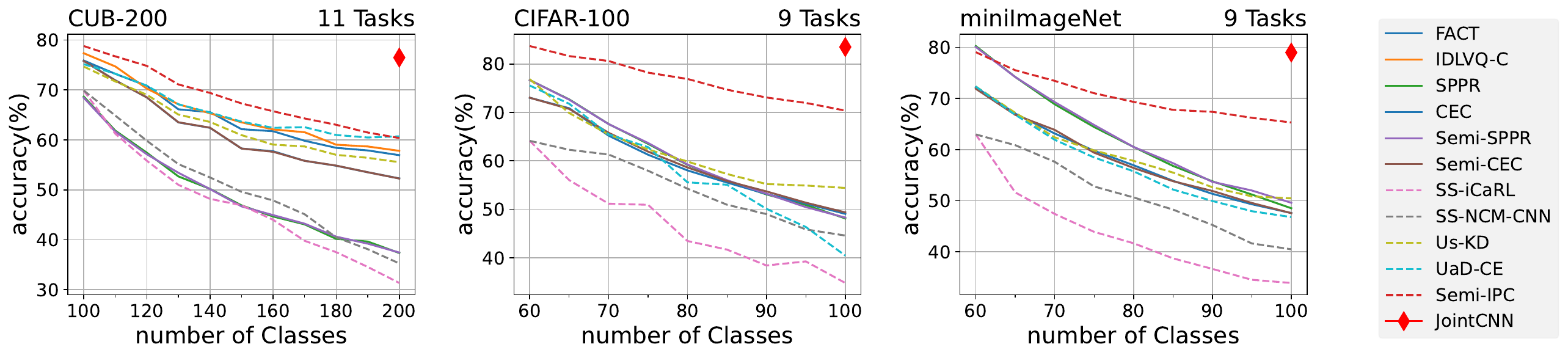}}
		\vskip -0.07in
		\caption{Comparisons of FSCIL and Semi-CIL methods on CUB-200, CIFAR-100, and miniImageNet. Dashed lines indicate non-exemplar methods, while solid lines represent exemplar-based methods. 
CUB-200 has a few additional unlabeled data ($|\mathcal{D}_{u}|=25$), and our method significantly outperforms SOTA on CIFAR-100, miniImageNet.}
		\label{figure-10}
	\end{center}
	\vskip -0.2in
\end{figure*}
\begin{table}[t]
	\caption{Average accuacies (\%) on CIFAR-100 and ImageNet-100. Exemplar-based CIL methods save 20/class samples.}
	\vskip -0.13in
	\label{table-3}
	\begin{center}
		\renewcommand\tabcolsep{2.9pt}
		\newcommand{\tabincell}[2]{\begin{tabular}{@{}#1@{}}#2\end{tabular}}
		\scalebox{1}{
			\renewcommand{\arraystretch}{1.25}
			\begin{tabular}{lcccccc}
				\toprule[1.3pt]
				\multirow{2}*{\tabincell{c}{\textbf{Method}}} &	
				\multicolumn{3}{c}{\makecell{CIFAR-100}} & \multicolumn{3}{c}{\makecell{ImageNet-100}}\\
				\cmidrule(lr){2-4} \cmidrule(lr){5-7}
				& $T=5$ &$T=10$ & $T=25$ & $T=5$ &$T=10$ & $T=25$\\
				\midrule
				iCaRL \cite{Rebuffi2017iCaRLIC}& 57.12 &52.66 &48.22 &65.44 &59.88 &52.97 \\
				LUCIR \cite{Rebuffi2017iCaRLIC}& 63.17&60.14&57.54&70.84&68.32&61.44\\
				PODNet \cite{Douillard2020SmallTaskIL}&64.83&63.19&60.72&75.54&74.33&68.31\\
				GeoDL \cite{simon2021learning}&65.14 &65.03 &63.12 &73.87 &73.55 &71.72 \\
				Mnemonics \cite{liu2020mnemonics}  &63.34&62.28&60.96&72.58&71.37&69.74\\
				AANets \cite{liu2021adaptive}  &66.31 &64.31 &62.31 &76.96 &75.58 &71.78\\
				CwD \cite{shi2022mimicking} &67.44&64.64&62.24&76.91&74.34&67.42\\
				SSIL \cite{ahn2021ss} & 63.02 &61.52 &58.02 &-- &-- &--\\
				DMIL \cite{tang2022learning}&68.01  &66.47 &--   &77.20 &\underline{76.76} &--\\
				AFC \cite{kang2022class}&66.49  &64.98 &63.89  &76.87 &75.75 &\underline{73.34}\\
				DualNet \cite{pham2021dualnet}&68.01 &63.42 &63.22&71.36&67.21&66.35\\
				BiMeCo \cite{nie2023bilateral} &\underline{69.87}&\underline{66.82}&\underline{64.16}&72.87&69.91&67.85\\
				EOPC \cite{wen2023optimizing} &67.55 &65.54 &61.82 &\textbf{78.95} &74.99 &70.10\\
				\cellcolor{mygray}Semi-IPC &\cellcolor{mygray}\textbf{76.11} &\cellcolor{mygray}\textbf{75.66}   &\cellcolor{mygray}\textbf{74.39}  &\cellcolor{mygray}\underline{77.51} &\cellcolor{mygray}\textbf{77.03}  &\cellcolor{mygray}\textbf{75.94}\\
				\bottomrule[1.3pt]
		\end{tabular}}
	\end{center}
\end{table}

\begin{table}[t]
	\caption{Comparisons on CIFAR-100 and ImageNet-100 under 5 tasks. Baselines results are come from \cite{hu2023dense}.}
	\vskip -0.13in
	\label{table-4}
	\begin{center}
		\renewcommand\tabcolsep{6.6pt}
		\newcommand{\tabincell}[2]{\begin{tabular}{@{}#1@{}}#2\end{tabular}}
		\scalebox{1}{
			\renewcommand{\arraystretch}{1.25}
			\begin{tabular}{clcccc}
				\toprule[1.3pt]
				\multirow{2}*{\tabincell{c}{\textbf{Exemplar}}} &	
				\multirow{2}*{\tabincell{c}{\textbf{Method}}} &	
				\multicolumn{2}{c}{\makecell{CIFAR-100}} & \multicolumn{2}{c}{\makecell{ImageNet-100}}\\
				\cmidrule(lr){3-4} \cmidrule(lr){5-6}
				&& Avg  &Last  & Avg  &Last \\
				\midrule
				\multirow{4}*{\tabincell{c}{2000}}
				&DER \cite{yan2021dynamically}&71.69 &63.78  & \underline{76.90}  &\underline{70.40} \\
				&FOSTER \cite{wang2022foster}&{72.20}  &63.31 &75.85  &67.68 \\
				&Dytox \cite{douillard2022dytox}&71.55  &{64.06} &75.54  &68.84 \\
				&DNE \cite{hu2023dense}&\underline{74.86}  &\textbf{70.04} &\textbf{78.56}  &\textbf{73.58} \\
				\cellcolor{mygray} 0 
				&\cellcolor{mygray}Semi-IPC &\cellcolor{mygray}\textbf{76.11} &\cellcolor{mygray}\underline{68.83} &\cellcolor{mygray}{76.48} &\cellcolor{mygray}{69.19}  \\
				\bottomrule[1.3pt]
		\end{tabular}}
	\end{center}
	\vskip -0.1in
\end{table}

\noindent\textbf{Contrastive pre-trained model.}
In the main experiment, the contrastive pre-trained model used is BYOL \cite{grill2020bootstrap}, with ResNet50 \cite{he2016deep} as the backbone network. For a fair comparison, we employed the same pre-trained model for other compared CIL methods. Experiments with different contrastive learning techniques are detailed in Sec. \ref{sec: 5.4.2}.
To avoid category information leakage \cite{kimtheoretical} in the continual learning process, we excluded the classes involved in CIL phase from the pre-training phase. Specifically, category from ImageNet-100 \cite{deng2009imagenet}, CIFAR-100 \cite{krizhevsky2009learning}, miniImageNet \cite{vinyals2016matching} and CUB-200 \cite{AGEM} were removed from the ImageNet \cite{deng2009imagenet} dataset to form the respective pre-training datasets.
This distinguishes our method from previous pre-training based methods \cite{wang2022dualprompt, wang2022learning}.

\vspace*{5pt}
\noindent
\textbf{Evaluation protocol and metrics.} 
Performance is assessed using three metrics following the setting in \cite{cui2023uncertainty}: (1) \textbf{\emph{Last Accuracy}} ($a_{\text{last}}$): this is the top-1 accuracy across all classes at the final Semi-CIL tasks. (2) \textbf{\emph{Average Accuracy}}, defined as $A_{t} = \frac{1}{t}\sum_{i=1}^{t}a_{i}$, where each $a_{i}$ is the top-1 accuracy up to task $i$. (3) \textbf{\emph{Performance Dropping Rate (PD) (\%)}}, measuring the absolute accuracy reduction from the initial to the final task, which is calculated as $a_{\text{1}} - a_{\text{last}}$.

%
%

\vspace*{5pt}
\noindent
%
\textbf{Implementation details.} In our main experiment, the data is equally divided into $T$ tasks (i.e., $\bm{B}=0$), following the configuration in \cite{yan2021dynamically}.
Moreover, we conduct experiments with a majority of classes assigned to the first task, and the rest is evenly allocated among other tasks (e.g., $\bm{B}=50,60,100$).
The networks are trained using the SGD optimizer with an initial learning rate of 0.1 and a momentum of 0.9.
We use the cosine learning rate decay \cite{loshchilov2016sgdr}. All models are trained with a batch size 128 for 80 epochs per phase.
For exemplar-based methods, the \textit{herd selection} \cite{rebuffi2017icarl} is used to select $\bm{R}$ samples for each old class.

\begin{table*}[t]
	\caption{Comparisons of task accuracies, average accuracies, and PD (\%) on CUB-200 under $B=100, T=11$ setup. Task 1 uses all data, FSCIL uses 5/class for tasks 2-12, and Semi-CIL additionally employs the remaining samples as unlabeled data. Baseline results come from \cite{cui2023uncertainty}.}
	\vskip -0.13in
	\label{table-5}
	\begin{center}
		\renewcommand\tabcolsep{4pt}
		\newcommand{\tabincell}[2]{\begin{tabular}{@{}#1@{}}#2\end{tabular}}
		\scalebox{1}{
			\renewcommand{\arraystretch}{1.3}
\begin{tabular}{cclccccccccccccc}
\toprule[1.3pt]
                             &                                     &                                   & \multicolumn{11}{c}{\textbf{Tasks}}                                                                                                                                                      &                       &                      \\ \cline{4-14}
\multirow{-2}{*}{\textbf{$\bm{|D_{u}|}$}} & \multirow{-2}{*}{\textbf{Exemplar}} & \multirow{-2}{*}{\textbf{Method}} & 1              & 2              & 3              & 4              & 5              & 6              & 7              & 8              & 9              & 10             & 11             & \multirow{-2}{*}{\textbf{Avg ($\uparrow$)}} & \multirow{-2}{*}{\textbf{PD ($\downarrow$)}} \\ \hline
                             &                                     & FACT \cite{zhou2022forward}                             & 75.90          & 73.23          & 70.84          & 66.13          & \underline{65.56}          & 62.15          & 61.74          & 59.83          & 58.41          & 57.89          & 56.94          & 64.42                 & 18.96                \\
                             &                                     & IDLVQ-C \cite{chen2020incremental}                          & \underline{77.37}          & \underline{74.72}          & 70.28          & 67.13          & 65.34          & 63.52          & 62.10          & 61.54          & 59.04          & 58.68          & 57.81          & 65.23                 & 19.56                \\
                             &                                     & SPPR \cite{zhu2021self}                             & 68.68          & 61.85          & 57.43          & 52.68          & 50.19          & 46.88          & 44.65          & 43.07          & 40.17          & 39.63          & 37.33          & 49.32                 & 31.35                \\
\multirow{-4}{*}{0}          & \multirow{-4}{*}{0}                & CEC \cite{zhang2021few}                              & 75.85          & 71.94          & 68.50          & 63.50          & 62.43          & 58.27          & 57.73          & 55.81          & 54.83          & 53.52          & 52.28          & 61.33                 & 23.57                \\ \hline \hline
                             &                                     & Semi-SPPR \cite{cui2023uncertainty}                        & 68.44          & 61.66          & 57.11          & 53.41          & 50.15          & 46.68          & 44.93          & 43.21          & 40.61          & 39.21          & 37.43          & 49.35                 & 31.01                \\
                             &                                     & Semi-CEC \cite{cui2023uncertainty}                         & 75.82          & 71.91          & 68.52          & 63.53          & 62.45          & 58.27          & 57.62          & 55.81          & 54.85          & 53.52          & 52.26          & 61.32                 & 23.56                \\
                             &                                     & SS-iCaRL \cite{cui2021semi}                         & 69.89          & 61.24          & 55.81          & 50.99          & 48.18          & 46.91          & 43.99          & 39.78          & 37.50          & 34.54          & 31.33          & 47.29                 & 38.56                \\
                             &                                     & SS-NCM-CNN \cite{cui2021semi}                       & 69.89          & 64.87          & 59.82          & 55.14          & 52.48          & 49.60          & 47.87          & 45.10          & 40.47          & 38.10          & 35.25          & 50.78                 & 34.64                \\
                             &                                     & Us-KD \cite{cui2022uncertainty}                            & 74.69          & 71.71          & 69.04          & 65.08          & 63.60          & 60.96          & 59.06          & 58.68          & 57.01          & 56.41          & 55.54          & 62.89                 & 19.15                \\
\multirow{-6}{*}{25}       & \multirow{-6}{*}{2000}                & UaD-CE \cite{cui2023uncertainty}                           & 75.17          & 73.27          & \underline{70.87}          & \underline{67.14}          & 65.49          & \underline{63.66}          & \underline{62.42}          & \underline{62.55}          & \underline{60.99}          & \underline{60.48}          & \underline{60.72}          & \underline{65.71}                 & \textbf{14.45}                \\
\rowcolor{mygray} 
25                         & 0                                   & Semi-IPC                & \textbf{78.81} & \textbf{76.74} & \textbf{74.83} & \textbf{71.85} & \textbf{70.21} & \textbf{68.40} & \textbf{66.46} & \textbf{64.71} & \textbf{63.03} & \textbf{61.48} & \textbf{60.37} & \textbf{68.81}        & \underline{18.44}      
\\ \bottomrule[1.3pt]
\end{tabular}

}
	\end{center}
\end{table*}

\begin{table*}[t]
	\caption{Comparisons of task accuracies, average accuracies, and PD (\%) on CIFAR-100 under $B=60, T=9$ setup.}
	\vskip -0.13in
	\label{table-6}
	\begin{center}
		\renewcommand\tabcolsep{4pt}
		\newcommand{\tabincell}[2]{\begin{tabular}{@{}#1@{}}#2\end{tabular}}
		\scalebox{1}{
			\renewcommand{\arraystretch}{1.3}
\begin{tabular}{cclccccccccccc}
\toprule[1.3pt]
                             &                                     &                                   & \multicolumn{9}{c}{\textbf{Tasks}}                                                                                                                     &                       &                      \\ \cline{4-12}
\multirow{-2}{*}{\textbf{$\bm{|D_u|}$}} & \multirow{-2}{*}{\textbf{Exemplar}} & \multirow{-2}{*}{\textbf{Method}} & 1              & 2              & 3              & 4              & 5              & 6              & 7              & 8              & 9              & \multirow{-2}{*}{\textbf{Avg ($\uparrow$)}} & \multirow{-2}{*}{\textbf{PD ($\downarrow$)}} \\ \hline
                             &                                     & SPPR \cite{zhu2021self}                             & 76.68          & \underline{72.69}          & 67.61          & 63.52          & 59.18          & 55.82          & 53.08          & 50.89          & 48.12          & 60.84                 & 28.56                \\
                             &                                     & CEC \cite{zhang2021few}                              & 73.03          & 70.86          & 65.20          & 61.27          & 58.03          & 55.53          & 53.17          & 51.19          & 49.06          & 59.70                 & 23.97                \\
\multirow{-3}{*}{0}          & \multirow{-3}{*}{0}                & UaD-CE \cite{cui2023uncertainty}                           & 75.55          & 71.78          & 65.47          & 62.83          & 55.56          & 55.08          & 50.11          & 46.35          & 40.46          & 58.13                 & 35.09                \\ \hline \hline
                             &                                     & Semi-SPPR \cite{cui2023uncertainty}                        & 76.68          & 72.63          & 67.59          & 63.69          & 59.24          & 56.02          & 53.23          & 50.46          & 48.29          & 60.87                 & 28.39                \\
                             &                                     & Semi-CEC \cite{cui2023uncertainty}                         & 73.03          & 70.72          & 65.79          & 61.91          & 58.64          & 55.84          & 53.70          & 51.37          & 49.37          & 60.04                 & 23.66                \\
                             &                                     & SS-iCaRL \cite{cui2021semi}                         & 64.13          & 56.02          & 51.16          & 50.93          & 43.46          & 41.69          & 38.41          & 39.25          & 34.80          & 46.65                 & 29.33                \\
                             &                                     & SS-NCM-CNN \cite{cui2021semi}                       & 64.13          & 62.29          & 61.31          & 57.96          & 54.26          & 50.95          & 49.02          & 45.85          & 44.59          & 54.48                 & 19.54                \\
                             &                                     & Us-KD \cite{cui2022uncertainty}                            & \underline{76.85}          & 69.87          & 65.46          & 62.36          & 59.86          & 57.29          & 55.22          & 54.91          & 54.42          & 61.80                 & 22.43                \\
\multirow{-6}{*}{495}        & \multirow{-6}{*}{2000}                & UaD-CE \cite{cui2023uncertainty}                           & 75.55          & 72.17          & \underline{68.57}          & \underline{65.35}          & \underline{62.80}          & \underline{60.27}          & \underline{59.12}          & \underline{57.05}          & \underline{54.50}          & \underline{63.93}                 & \underline{21.05}                \\
\rowcolor{mygray} 
495                          & 0                                   & Semi-IPC                          & \textbf{83.70} & \textbf{81.63} & \textbf{80.61} & \textbf{78.21} & \textbf{76.91} & \textbf{74.70} & \textbf{73.08} & \textbf{71.96} & \textbf{70.38} & \textbf{76.80}        & \textbf{13.32}      
\\ \bottomrule[1.3pt]
\end{tabular}
}
	\end{center}
\end{table*}

\begin{table*}[t]
	\caption{Comparisons of task accuracies, average accuracies, and PD (\%) on miniImageNet under $B=60, T=9$ setup.}
	\vskip -0.13in
	\label{table-7}
	\begin{center}
		\renewcommand\tabcolsep{4pt}
		\newcommand{\tabincell}[2]{\begin{tabular}{@{}#1@{}}#2\end{tabular}}
		\scalebox{1}{
			\renewcommand{\arraystretch}{1.3}
\begin{tabular}{cclccccccccccc}
\toprule[1.3pt]
                             &                                     &                                   & \multicolumn{9}{c}{\textbf{Tasks}}                                                                                                                     &                       &                      \\ \cline{4-12}
\multirow{-2}{*}{\textbf{$\bm{|D_u|}$}} & \multirow{-2}{*}{\textbf{Exemplar}} & \multirow{-2}{*}{\textbf{Method}} & 1              & 2              & 3              & 4              & 5              & 6              & 7              & 8              & 9              & \multirow{-2}{*}{\textbf{Avg ($\uparrow$)}} & \multirow{-2}{*}{\textbf{PD ($\downarrow$)}} \\ \hline
                             &                                     & SPPR \cite{zhu2021self}                             & \underline{80.27}          & \underline{74.22}          & \underline{68.89}          & 64.43          & \underline{60.54}          & \underline{56.82}          & \underline{53.81}          & 51.22          & 48.54          & \underline{62.08}                 & 31.73                \\
                             &                                     & CEC \cite{zhang2021few}                              & 72.22          & 67.06          & 63.17          & 59.79          & 56.96          & 53.91          & 51.36          & 49.32          & 47.60          & 57.93                 & 24.62                \\
\multirow{-3}{*}{0}          & \multirow{-3}{*}{0}                 & UaD-CE \cite{cui2023uncertainty}                           & 72.35          & 66.83          & 61.94          & 58.48          & 55.77          & 52.20          & 49.96          & 47.96          & 46.81          & 56.92                 & 25.54                \\ \hline \hline
                             &                                     & Semi-SPPR \cite{cui2023uncertainty}                        & 80.10          & 74.21          & 69.31          & \underline{64.83}          & 60.53          & 57.36          & 53.70          & \underline{52.01}          & 49.61          & 62.41                 & 30.49                \\
                             &                                     & Semi-CEC \cite{cui2023uncertainty}                         & 71.91          & 66.81          & 63.87          & 59.41          & 56.42          & 53.83          & 51.92          & 49.57          & 47.58          & 57.92                 & 24.33                \\
                             &                                     & SS-iCaRL \cite{cui2021semi}                         & 62.98          & 51.64          & 47.43          & 43.92          & 41.69          & 38.74          & 36.67          & 34.54          & 33.92          & 43.50                 & 29.06                \\
                             &                                     & SS-NCM-CNN \cite{cui2021semi}                       & 62.98          & 60.88          & 57.63          & 52.80          & 50.66          & 48.28          & 45.27          & 41.65          & 40.51          & 51.18                 & 22.47                \\
                             &                                     & Us-KD \cite{cui2022uncertainty}                            & 72.35          & 67.22          & 62.41          & 59.85          & 57.81          & 55.52          & 52.64          & 50.86          & \underline{50.47}          & 58.79                 & 21.88                \\
\multirow{-6}{*}{595}        & \multirow{-6}{*}{2000}              & UaD-CE \cite{cui2023uncertainty}                           & 72.35          & 66.91          & 62.13          & 59.89          & 57.41          & 55.52          & 53.26          & 51.46          & 50.52          & 58.83                 & \underline{21.83}                \\
\rowcolor{mygray} 
595                          & 0                                   & Semi-IPC                          & \textbf{79.07} & \textbf{75.54} & \textbf{73.46} & \textbf{71.04} & \textbf{69.35} & \textbf{67.79} & \textbf{67.40} & \textbf{66.23} & \textbf{65.32} & \textbf{70.58}        & \textbf{13.75}      
\\ \bottomrule[1.3pt]
\end{tabular}

}
	\end{center}
\end{table*}

\subsection{Comparative results}
\noindent\textbf{Comparison with standard CIL methods.} We compare with CIL methods, including exemplar-based and non-exemplar methods. 
Employing full category annotations and contrastive pre-trained models, these methods provide a solid baseline for evaluation.
Among them, DER \cite{yan2021dynamically}, PASS \cite{Zhu_2021_CVPR}, and FeTrIL \cite{Petit_2023_WACV} are SOTA methods. 
From Table~\ref{table-2}, Fig.~\ref{figure-8}, and Fig.~\ref{figure-9}, it is observed that existing CIL methods have two limitations. First, they rely on preserving old samples because non-exemplar methods typically perform worse than exemplar-based methods. Second, when annotations are reduced (e.g., ImageNet-100 $\Rightarrow$ miniImageNet), the average accuracy of DER decreases from 79.12\% to 66.26\%.
Without storing old examples and leveraging limited labeled data, our method performs comparable or better than DER, e.g., Semi-IPC surpasses DER by 10.86\% in the 10-tasks setting on miniImageNet. Note that DER continuously expands the feature extractor, whereas our method utilizes a single, fixed feature extractor.
Compared to non-exemplar approaches, our method shows more significant improvement, outperforming SOTA methods by large margins.
Tables~\ref{table-3} and~\ref{table-4} compare Semi-IPC with latest exemplar-based methods under the settings of $R=20$ and $B=50$, where most results are from their original papers. We can observe that our method consistently outperforms strong baselines like CwD \cite{shi2022mimicking}, DMIL \cite{tang2022learning}, AFC \cite{kang2022class}, EOPC \cite{zhu2023imitating}, etc. Particularly, our method achieves similar results with  Dytox \cite{douillard2022dytox} and DNE \cite{hu2023dense}, which leverage more advanced transformer architecture \cite{dosovitskiyimage}.
These results demonstrate that even without storing any old examples, an incremental learner can indeed perform well by leveraging a large amount of unlabeled data.
\vspace*{5pt}


\vspace*{5pt}
\noindent\textbf{Comparison with FSCIL methods.} 
FSCIL is a label-efficient paradigm that only uses a few (e.g., 5 per class) labeled samples to learn new classes. 
As shown in Tables~\ref{table-5},~\ref{table-6},~\ref{table-7} and Fig.~\ref{figure-8}, leveraging easily obtained unlabeled data, our method achieved a performance improvement up to 18.67\% on the CIFAR-100 and miniImageNet compared to methods like FACT \cite{zhou2022forward}, CEC \cite{zhang2021few}, etc. 
Additionally, under the FSCIL setting, new classes suffer from overfitting problems due to the limited labeled example, leading to a significant imbalance between old and new classes. Our method demonstrates a more balanced recognition capability, which is detailed in Sec ~\ref{sec: 5.3.1}.
\vspace*{5pt}
\noindent\textbf{Comparison with Semi-FSCIL methods.} Recently, some research incorporated unsupervised learning into the FSCIL setting, such as SS-iCaRL \cite{cui2021semi}, SS-NCM-CNN \cite{cui2021semi}, Us-KD Us-KD \cite{cui2022uncertainty}, and UaD-CE \cite{cui2023uncertainty}. Besides, some FSCIL methods can be easily modified \cite{cui2023uncertainty}, like Semi-CEC and Semi-SPPR.
These methods are denoted as Semi-FSCIL.
As shown in Tables~\ref{table-5},~\ref{table-6},~\ref{table-7}, and Fig.~\ref{figure-8}, we find that our method significantly outperforms recently developed semi-FSCIL methods. Among them, UaD-CE is currently the most advanced, retaining 20 samples per category. Despite our method being non-exemplar, it surpasses UaD-CE in terms of average accuracy and PD metric across three benchmark datasets. For instance, on the CIFAR-100 dataset, UaD-CE achieves an average accuracy of 65.71\%, while our method reaches 76.80\%. This demonstrates our superior performance in more realistic and challenging settings.

\subsection{Further Analysis}  
\subsubsection{Balancing base and novel classes}
\label{sec: 5.3.1}
As shown in Table \ref{table-8}, FSCIL methods typically showed poor balance between new and old classes, which mainly stemmed from two factors: 
\textbf{(1)} Unequal data division, with an excessive proportion in task 1; 
\textbf{(2)} Imbalance in classifier and feature extractor during continual learning. 
The dynamic architectures exemplar-based method (e.g., DER) performed best, achieving an average accuracy of 75.9\% for old classes and 66.39\% for new. This is attributed to dynamic architectures maintaining multiple feature extractors and saving old data mitigating classifier bias.
Benefiting from the integration of contrastive learning with Semi-IPC for enhancing the balance in feature extraction and classifier among new and old classes, our method achieved average accuracy rates of 65.66\% for new classes and 80.21\% for old ones. It was second only to DER, outperforming SOTA methods like FeTrIL and UaD-CE, which rely on complete annotation information and save 20 old samples per class, respectively. Based on these results, our non-exemplar Semi-CIL method achieved better balance, effectively mitigated overfitting on emerging categories, and maintained robust performance across learned categories.

\subsubsection{Robust to OOD unlabelled data}
We conducted experiments to test the performance of our model with out-of-distribution (OOD) unlabelled data from class-irrelevant datasets of ImageNet. We resized the images for CUB-200, miniImageNet, and CIFAR-100. To evaluate the impact of adding unknown class data to the original in-distribution datasets, we added 5\%-20\% OOD unlabelled data and measured the average accuracy in the UaD-CE setting.
As demonstrated in Table \ref{table-ood}, the state-of-the-art method UaD-CE is relatively sensitive to OOD data. For instance, a 20\% increase in OOD data led to a 4.50\% decrease in accuracy on the CIFAR-100 dataset, with similar results observed on miniImageNet and CUB-200.
Using prototype classifiers, our model effectively rejects class-irrelevant unlabeled data based on Eq.~(\ref{eq16}), significantly reducing the impact of OOD data. On CIFAR-100, the model's average accuracy varied only by 0.10\%-0.32\%. This indicates that our approach to utilizing unlabeled samples is more robust, suitable for real-world challenges, and capable of handling in-distribution samples from known categories as well as out-of-distribution samples from unknown categories in the unlabeled data.

\begin{table}[t]
	\caption{Comparisons of average and last incremental accuracies (\%) on CIFAR-100 of base and novel classes. Exemplar-based methods save 20/class samples. Baselines results are come from \cite{cui2023uncertainty}.}
	\vskip -0.13in
	\label{table-8}
	\begin{center}
		\renewcommand\tabcolsep{3.8pt}
		\newcommand{\tabincell}[2]{\begin{tabular}{@{}#1@{}}#2\end{tabular}}
		\scalebox{1}{
			\renewcommand{\arraystretch}{1.3}
\begin{tabular}{lccccccc}
\toprule[1.3pt]
                                  & \multicolumn{2}{c}{\textbf{Base}} & \multicolumn{2}{c}{\textbf{Novel}} & \multicolumn{2}{c}{\textbf{All}} &                                 \\
                                  \cmidrule(lr){2-3} \cmidrule(lr){4-5}\cmidrule(lr){6-7}  
\multirow{-2}{*}{\textbf{Method}} & Avg             & Last            & Avg              & Last            & Avg             & Last           & \multirow{-2}{*}{\textbf{Diff ($\downarrow$)}} \\ \hline
DER \cite{yan2021dynamically}                              & 75.90           & 65.62           & \textbf{66.39}            & \textbf{60.75}           & \underline{74.17}           & 63.67          & \textbf{9.51}                            \\
FeTrIL \cite{Petit_2023_WACV}                           & \underline{77.46}           & 73.92           & 62.58            & 52.30           & 73.52           & \underline{65.27}          & 14.88                           \\
\hline
SPPR \cite{zhu2021self}                             & 74.69           & \underline{75.30}           & 7.28             & 7.35            & 60.84           & 48.12          & 67.41                           \\
CEC \cite{zhang2021few}                              & 70.38           & 67.78           & 24.38            & 20.98           & 59.70           & 49.06          & 46.00                           \\
\hline
SS-iCaRL \cite{cui2021semi}                         & 50.75           & 40.97           & 31.01            & 26.10           & 46.65           & 34.80          & 19.74                           \\
SS-NCM \cite{cui2021semi}                       & 59.48           & 52.49           & 37.16            & 33.44           & 54.48           & 44.59          & 22.32                           \\
UaD-CE \cite{cui2023uncertainty}                           & 68.36           & 61.99           & 49.74            & 42.27           & 63.93           & 54.50          & 18.62                           \\
\rowcolor{mygray} 
Semi-IPC                          & \textbf{80.21}           & \textbf{76.90}           & \underline{65.66}            & \underline{60.60}           & \textbf{76.80}           & \textbf{70.38}          & \underline{14.55}                           
\\ \bottomrule[1.3pt]
\end{tabular}
		}
	\end{center}
 \vskip -0.08in
\end{table}

\begin{table}[t]

	\caption{Comparisons of average accuracies (\%) on CIFAR-100, miniImageNet, and CUB-200 with adding unlabelled data from unknown categories, following the setting in \cite{cui2023uncertainty}.}
 	\vskip -0.13in
	\label{table-ood}
	\begin{center}
		\renewcommand\tabcolsep{6pt}
		\newcommand{\tabincell}[2]{\begin{tabular}{@{}#1@{}}#2\end{tabular}}
		\scalebox{1}{
			\renewcommand{\arraystretch}{1.3}
\begin{tabular}{clcccc}
				\toprule[1.3pt]
                                    &                                   & \multicolumn{4}{c}{\textbf{Unknow Classes Data}}                                                                                                                    \\ \cline{3-6} 
\multirow{-2}{*}{\textbf{Datasets}} & \multirow{-2}{*}{\textbf{Method}} & 0\%                                    & 5\%                           & 10\%                          & 20\%                                                    \\ \hline
                                    & UaD-CE \cite{cui2023uncertainty}                            & 63.93                                  & 62.74                         & 62.11                         & 60.30                                                  \\
\multirow{-2}{*}{CIFAR-100}         & \cellcolor{mygray}Semi-IPC  & \cellcolor{mygray}\textbf{76.80} & \cellcolor{mygray}76.48 & \cellcolor{mygray}\underline{76.70} & \cellcolor{mygray}76.61 \\
                                    & UaD-CE \cite{cui2023uncertainty}                            & 58.82                                  & 56.92                         & 55.70                         & 55.11                                                \\
\multirow{-2}{*}{miniImageNet}      & \cellcolor{mygray}Semi-IPC  & \cellcolor{mygray}\textbf{70.58} & \cellcolor{mygray}\underline{70.43} & \cellcolor{mygray}70.40 & \cellcolor{mygray}70.35  \\
                                    & UaD-CE \cite{cui2023uncertainty}                           & 65.70                                  & -                             & 64.27                         & 63.60                                                 \\
\multirow{-2}{*}{CUB-200}           & \cellcolor{mygray}Semi-IPC  & \cellcolor{mygray}\textbf{68.81} & \cellcolor{mygray}-     & \cellcolor{mygray}\underline{68.68} & \cellcolor{mygray}68.50     \\ \bottomrule[1.3pt]
		\end{tabular} }
	\end{center}
	\vskip -0.1in
\end{table}

\begin{table*}[t]
	\caption{The effectiveness of each component in Semi-IPC. The baseline is classified using class means.}
	\vskip -0.13in
	\label{table-10}
	\begin{center}
		\renewcommand\tabcolsep{6pt} 
		\newcommand{\tabincell}[2]{\begin{tabular}{@{}#1@{}}#2\end{tabular}}
		\scalebox{1}{
			\renewcommand{\arraystretch}{1.23}
			\begin{tabular}{lcccccccccccc}
				\toprule[1.3pt]

\multirow{3}{*}{\textbf{Method}} & \multicolumn{4}{c}{\textbf{CIFAR-100}}                            & \multicolumn{4}{c}{\textbf{ImageNet-100}}                         & \multicolumn{4}{c}{\textbf{miniImageNet}}                         \\ \cmidrule(lr){2-5} \cmidrule(lr){6-9} \cmidrule(lr){10-13}	
                                 & \multicolumn{2}{c}{T=5}         & \multicolumn{2}{c}{T=10}        & \multicolumn{2}{c}{T=5}         & \multicolumn{2}{c}{T=10}        & \multicolumn{2}{c}{T=5}         & \multicolumn{2}{c}{T=10}        \\ \cmidrule(lr){2-3} \cmidrule(lr){4-5} \cmidrule(lr){6-7} \cmidrule(lr){8-9} \cmidrule(lr){10-11} \cmidrule(lr){12-13}	
                                 & Avg            & Last           & Avg            & Last           & Avg            & Last           & Avg            & Last           & Avg            & Last           & Avg            & Last           \\ \midrule
Baseline                         & 63.21          & 54.67          & 63.18          & 52.56          & 67.22          & 60.66          & 69.93          & 61.76          & 62.16          & 53.98          & 62.24          & 51.96          \\
+ Prototype                      & 64.83          & 56.92          & 65.91          & 54.88          & 67.33          & 54.92          & 65.84          & 51.96          & 64.15          & 53.44          & 63.44          & 51.32          \\
+ PUR                   & 71.78          & 62.28          & 71.21          & 60.05          & 68.57          & 56.18          & 67.20          & 54.56          & 67.40          & 57.54          & 66.31          & 55.38          \\
+ IU                             & 73.99          & 65.49          & 72.61          & 62.02          & 69.32          & 57.88          & 69.39          & 56.78          & 67.93          & 57.66          & 67.30          & 56.54          \\
+ PL Loss                        & \underline{75.28}          & \underline{66.70}          & \underline{74.23}          & \underline{64.22}          & \underline{74.82}          & \underline{66.68}          & \underline{75.31}          & \underline{66.46}          & \underline{68.47}          & \underline{59.81}          & \underline{70.25}          & \underline{60.18}          \\
+ Resample                       & \textbf{75.81} & \textbf{67.64} & \textbf{74.92} & \textbf{65.29} & \textbf{76.00} & \textbf{68.04} & \textbf{75.52} & \textbf{67.02} & \textbf{68.75} & \textbf{60.70} & \textbf{70.66} & \textbf{60.54}    \\ 
				\bottomrule[1.3pt]
		\end{tabular}}
	\end{center}
\end{table*}

\begin{figure*}[h]
	\begin{center}
		\centerline{\includegraphics[width=0.85\textwidth]{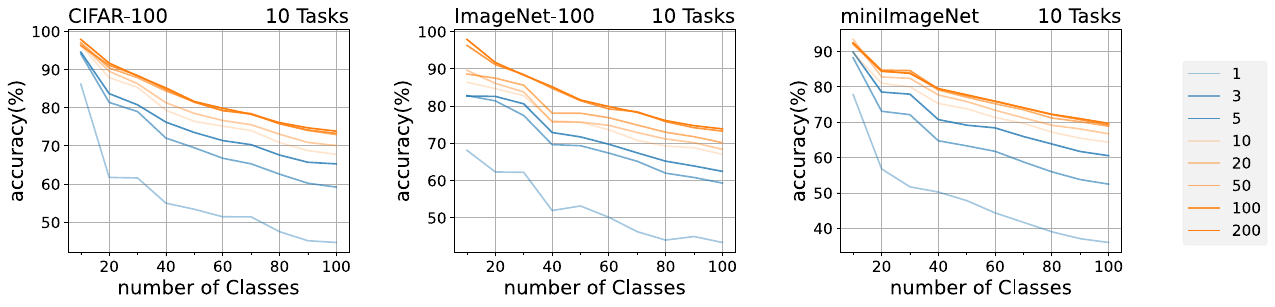}}
		\vskip -0.07in
		\caption{Analysis of varying annotation quantities under $B=0, T=10$ setting. In low-label situations, increasing labeled data significantly improves performance. With 100 to 200 labeled samples, our method nearly matches full-label performance.}
		\label{figure-12}
	\end{center}
	\vskip -0.2in
\end{figure*}

\begin{table}[t]
	\caption{The effectiveness of different contrastive learning strategies on CIFAR-100.}
	\vskip -0.13in
	\label{table-11}
	\begin{center}
		\renewcommand\tabcolsep{7pt}
		\newcommand{\tabincell}[2]{\begin{tabular}{@{}#1@{}}#2\end{tabular}}
		\scalebox{1}{
			\renewcommand{\arraystretch}{1.23}
\begin{tabular}{lcccc}
\toprule[1.3pt]

                                  & \multicolumn{2}{c}{\textbf{T=5}} & \multicolumn{2}{c}{\textbf{T=10}} \\
\cmidrule(lr){2-3} \cmidrule(lr){4-5}
\multirow{-2}{*}{\textbf{Method}} & Avg             & Last           & Avg             & Last            \\
\midrule
Supervised \cite{he2016deep}                       & 65.09           & 54.19          & 64.43           & 52.70           \\
MoCoV2 \cite{he2020momentum}                           & 69.57           & 59.14          & 68.84           & 57.49           \\
SimCLR \cite{chen2020simple}                           & \underline{72.01}           & 63.09          & \underline{71.85}           & \underline{61.36}           \\
SWAV \cite{caron2020unsupervised}                             & 69.19           & 59.80          & 68.52           & 57.00           \\
Barlow \cite{zbontar2021barlow}                           & 71.99           & \underline{63.74}          & 71.84           & 60.59           \\
SimSiam \cite{chen2021exploring}                          & 68.98           & 60.62          & 67.92           & 56.91           \\
\rowcolor{mygray} 
BYOL \cite{grill2020bootstrap}                             & \textbf{75.81}  & \textbf{67.64} & \textbf{74.92}  & \textbf{65.29}              
\\ \bottomrule[1.3pt]
\end{tabular}
		}
	\end{center}
 \vskip -0.25in
\end{table}
\begin{figure}[t]
	\begin{center}
		\centerline{\includegraphics[width=1.0\columnwidth]{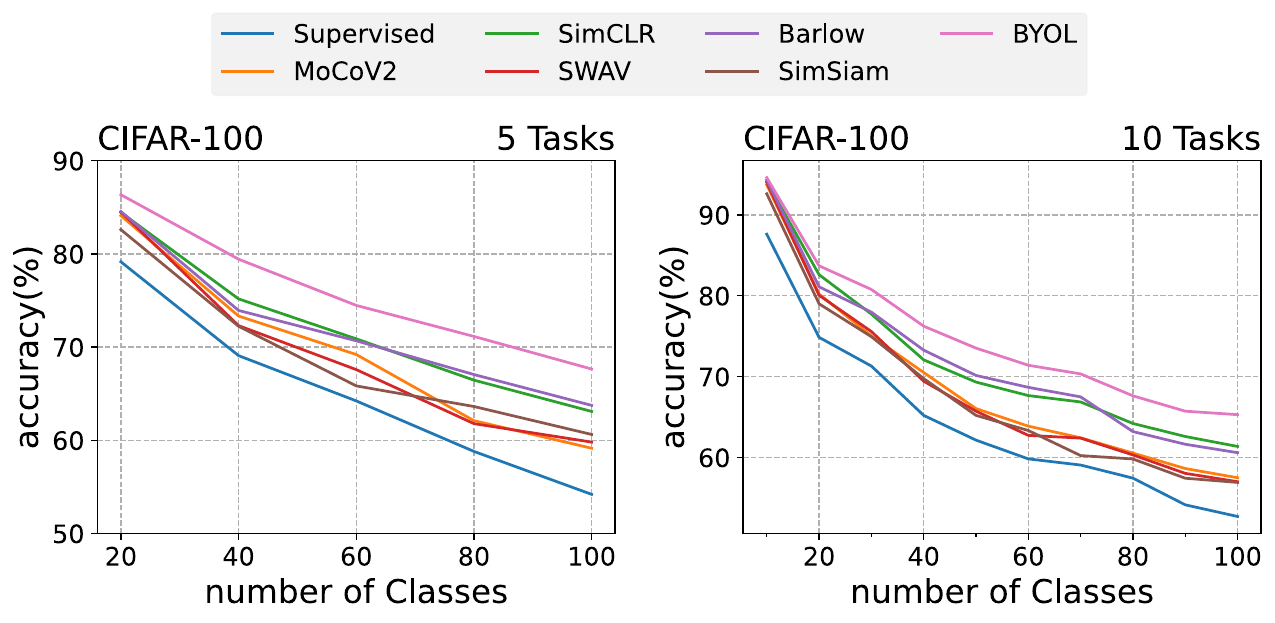}}
		\vskip -0.07in
		\caption{Accuracy curves under 5/10 task settings on CIFAR-100 when using different contrastive learning strategies.}
		\label{figure-13}
	\end{center}
	\vskip -0.32in
\end{figure}

\subsection{Ablation study}
\subsubsection{Impact of components in Semi-IPC.}
In the ablation study, we analyzed the impact of each component in Semi-IPC. The baseline model is the nearest-mean-of-examples classifier (NME), and IU is the gradient clipping incremental update strategy ( Sec. \ref{sec: 4.2.1} ). Results from Table \ref{table-10} show:
\textbf{(1)} Prototype learning improved accuracy over NME but caused more severe forgetting. This indicates that constraining prototypes during the incremental learning process and preventing forgetting are necessary.
\textbf{(2)} Prototype Unsupervised Regularization (PUR) of the classifier significantly improved model accuracy, with an average increase of 6.95\% in the 10-task setting. It demonstrates that our approach is effective in learning from unlabeled data.
\textbf{(3)} The combination of prototype learning loss (PL), IU, and prototype resample was effective in mitigating forgetting, increasing the last accuracy up to 12.46\%.
These results indicate that the combination of these components in Semi-IPC is crucial for balancing new class recognition with preventing old class forgetting.


\subsubsection{Impact of contrastive learning strategies.} 
\label{sec: 5.4.2}
We assessed various contrastive learning methods on Semi-IPC, including BYOL \cite{grill2020bootstrap}, MoCoV2 \cite{he2020momentum}, SimCLR \cite{chen2020simple}, SwAV \cite{caron2020unsupervised}, BarlowTwins \cite{zbontar2021barlow}, and SimSiam \cite{chen2021exploring}.
Additionally, we tested the supervised method, which pre-trained on ImageNet (excluding CIFAR-100 related classes). 
As analyzed in Sec. \ref{sec: 4.1}, contrastive learning acquired richer features, showing better generalization in new classes.
According to Table \ref{table-11} and Fig. \ref{figure-13}, Contrastive learning exhibited enhanced performance compared to supervised learning, showing an average accuracy improvement of up to 10.72\% and 10.49\% in the 5-step and 10-step tasks, respectively. 
Among these contrastive learning methods, BYOL proved to be the most suitable, primarily due to the use of Euclidean distance-based similarity measure, aligning with Semi-IPC.

\subsubsection{Impact of labeled data}
In our main experiments, each class had only 5 labeled samples, and the rest were unlabeled, a typical 5-shot setup in Semi-CIL and FSCIL (Tables \ref{table-1} and \ref{table-2}). 
Fig. \ref{figure-12} shows how varying the number of labeled data affects model performance:
\textbf{(1)} With only one labeled sample per class, the average accuracy was 55.82\% on ImageNet-100, comparable to iCaRL \cite{rebuffi2017icarl} (58.56\%). It reveals that our method remains effective even in situations where labels are extremely scarce, achieving acceptable CIL performance through learning from unlabeled data.
\textbf{(2)} Increasing labeled samples from 3 to 20 significantly enhanced performance. For instance, on CIFAR-100, average accuracy improved from 71.02\% to 79.87\%, and on ImageNet-100.
This shows our method is robust with varying label quantities, enhancing performance with more labeled data.
\textbf{(3)} Performance improvements were minor with more than 100 labeled samples.  
These results demonstrate that our model is effective with fewer labeled samples, significantly reducing dependence on labeled data.

\begin{figure}[t]
	\begin{center}
		\centerline{\includegraphics[width=0.9\columnwidth]{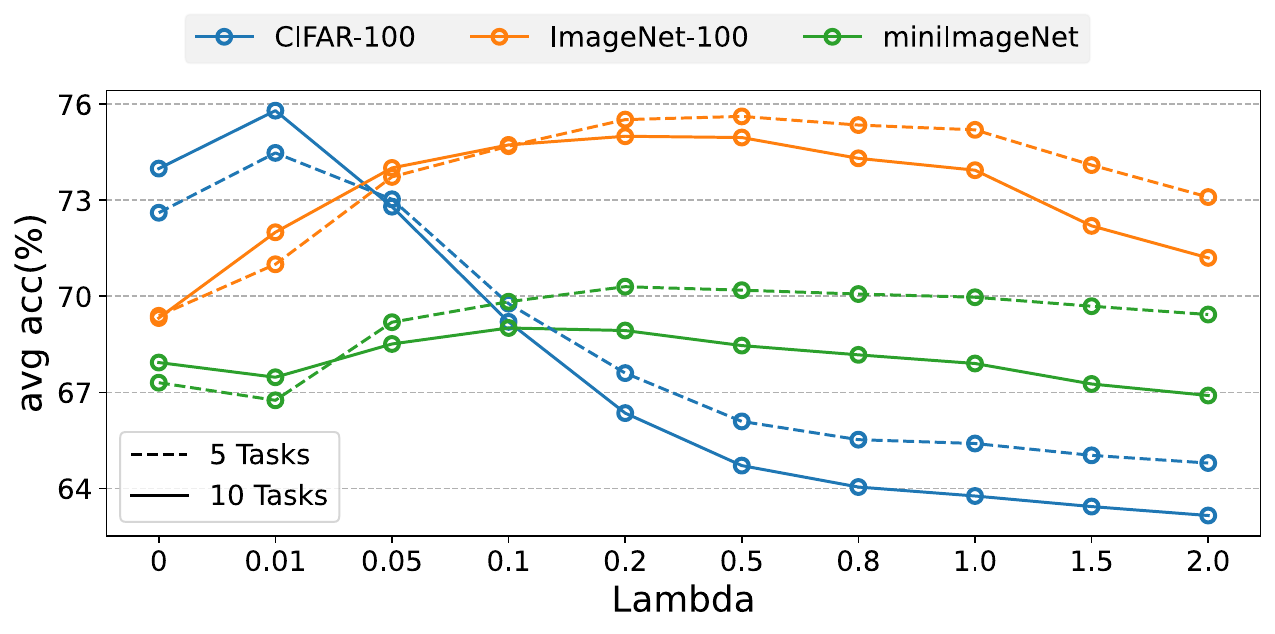}}
		\vskip -0.07in 
		\caption{Comparison of different lambda values on average accuracy ($B=0, T=5/10$). }
		\label{figure-14}
	\end{center}
	\vskip -0.25in
\end{figure}

\begin{figure}[t]
	\begin{center}
		\centerline{\includegraphics[width=0.95\columnwidth]{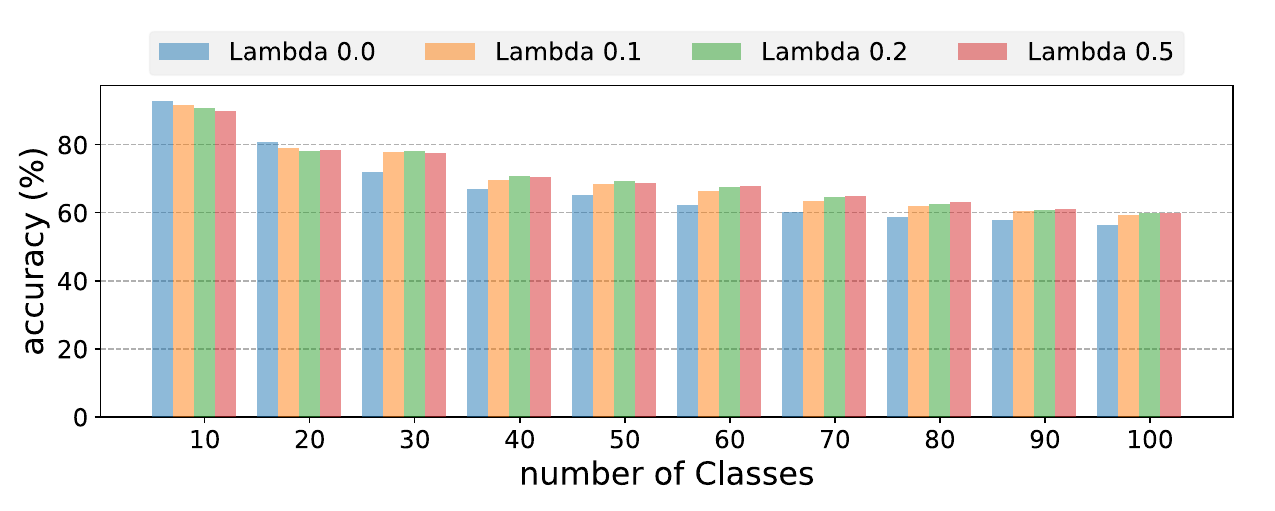}}
		\vskip -0.07in
		\caption{Comparisions of different lambda values on task accuracy in miniImageNet ($B=0, T=10$) .}
		\label{figure-15}
	\end{center}
	\vskip -0.25in
\end{figure}

\subsubsection{Impact of Hyperparameter.} 
Our method involves a single hyperparameter: the PL loss coefficient $\lambda$.
Since PL loss is calculated based on Euclidean distance, we adjust $\lambda$ to ensure the PL loss stays within a reasonable range. 
As shown in Fig. \ref{figure-14}, we tested the effect of different $\lambda$ values. 
A closer look at Fig. \ref{figure-15} reveals the impact of varying lambda values on task accuracies. 
A higher lambda value intensifies the PL loss constraint, resulting in lower accuracy in initial tasks but higher accuracy in subsequent tasks, suggesting a decrease in the model's forgetting.
Those results demonstrate that PL loss can effectively balance the accuracy of new and old classes.

\subsubsection{Comparison with semi-supervised learning methods}
\begin{figure}[t]
	\begin{center}
		\centerline{\includegraphics[width=0.8\columnwidth]{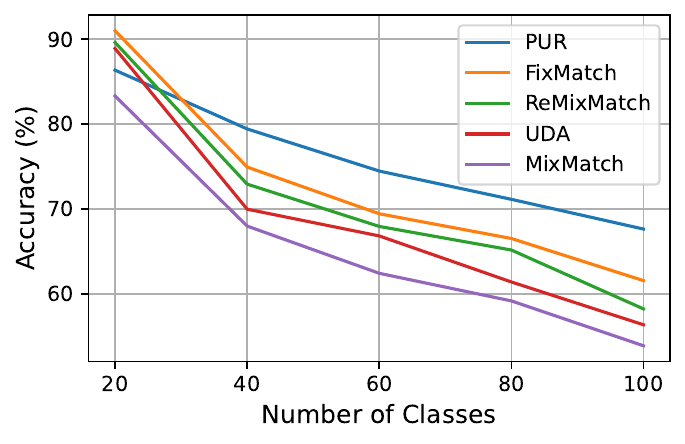}}
		\vskip -0.07in
		\caption{Comparision of different semi-supervised learning methods on task accuracy in CIFAR-100 ($B=0, T=5$) .}
		\label{figure-semiab}
	\end{center}
	\vskip -0.25in
\end{figure}

Prototype Unsupervised Regularization (PUR) is used in our method to learn from unlabeled data. We initialized representative semi-supervised methods with BYOL and applied them to Semi-CIL, comparing them with PUR, such as UDA \cite{xie2020unsupervised}, MixMatch \cite{berthelot2019mixmatch}, FixMatch \cite{berthelot2019remixmatch}, and ReMixMatch \cite{sohn2020fixmatch}. Results shown in Fig. \ref{figure-semiab} reveal that although FixMatch and ReMixMatch achieved higher accuracy on initial tasks, PUR surpassed them as the number of incremental tasks increased. This is because, in contrast to standard semi-supervised learning, Semi-CIL involves learning new categories while ensuring existing knowledge is not forgotten. This increases the complexity of applying these methods in continual learning settings, aligning with insights from \cite{cui2023uncertainty}.

\section{Conclusion}
\label{sec: 6}
In this paper, we investigate a challenging yet practical incremental learning scenario, i.e., non-exemplar semi-supervised class-incremental learning, where the model only has access to new classes with a few labeled and many unlabeled training examples. We propose a simple and effective framework with contrastive learning and semi-supervised incremental prototype classifier. Extensive experiments on benchmark datasets demonstrate that our method can achieve state-of-the-art performance without storing any old samples and only using less than 1\% of labeled samples in each incremental stage. 
Moreover, the proposed framework can be integrated with various CIL and FSCIL methods, facilitating the exploration of enhancements in semi-supervised and incremental learning methods.
In the future, we will consider adapting feature representations from the continuously increasing unlabelled data, and the latest developments in continual self-supervised learning \cite{fini2022self, marsocci2022continual} will be leveraged for valuable guidance.




%

%
%
%
%
%
%

\ifCLASSOPTIONcaptionsoff
  \newpage
\fi



%
\bibliography{refer}
\bibliographystyle{unsrt}
\end{document}